\documentclass[10pt,letterpaper]{article}
\ifdefined\pdfminorversion\pdfminorversion=7\fi
\usepackage[T1]{fontenc}
\usepackage[pagenumbers]{westlakeagi}
\usepackage{tcolorbox}
\usepackage[pagebackref,breaklinks,colorlinks,allcolors=wagiblue]{hyperref}

\usepackage{colortbl}
\usepackage{algorithm}
\usepackage{float}
\usepackage{algpseudocode}
\usepackage{arydshln}
\usepackage{placeins}
\usepackage{flafter}
\usepackage{tikz}
\usetikzlibrary{arrows.meta,calc,fit,positioning}

\newcommand{\fitwidth}[1]{%
  \resizebox{\ifdim\width>\linewidth \linewidth\else\width\fi}{!}{#1}}

\newcommand{\method}{Slot3R}

\newcommand{\papertitle}{\method: Set-Associative Spatial Memory for \\ Streaming 3D Reconstruction}
\title{\papertitle}
\author{Xiyuan Zhang, Yanming Yang, Kaiyuan Xu, Ruibo Li, Chi Zhang}
\date{}
\wagilogo{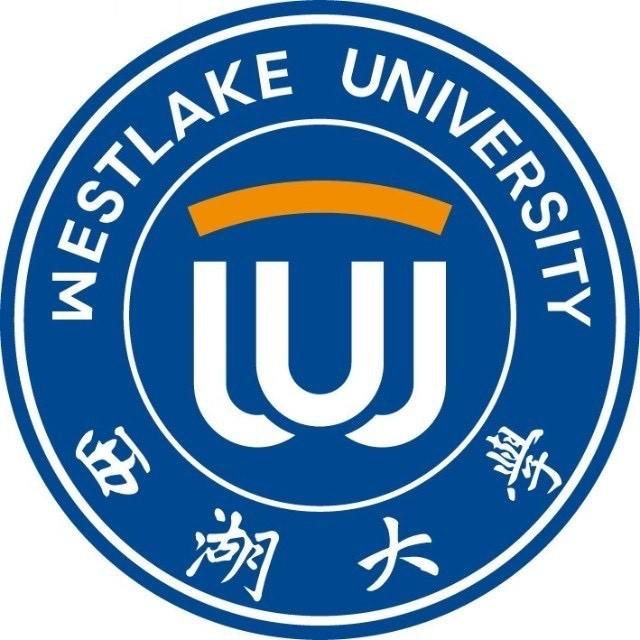}
\hypersetup{pdftitle={Slot3R: Set-Associative Spatial Memory for Streaming 3D Reconstruction},
  pdfauthor={Xiyuan Zhang, Yanming Yang, Kaiyuan Xu, Ruibo Li, Chi Zhang}}

\begin{document}

\vspace*{-0.42in}%
\wagibanner
\vskip 0.16in
\begin{tcolorbox}[colback=wagiblue!4, colframe=wagiblue!40,
                  boxrule=0.6pt, arc=10pt,
                  left=22pt, right=22pt, top=14pt, bottom=10pt]
  \begin{center}
    {\LARGE\bfseries \papertitle\par}
    \vskip 12pt
    {\large
      Xiyuan Zhang\textsuperscript{1,2*} \quad
      Yanming Yang\textsuperscript{1*} \quad
      Kaiyuan Xu\textsuperscript{1} \\
      Ruibo Li\textsuperscript{3} \quad
      Chi Zhang\textsuperscript{1$\dagger$}\par}
    \vskip 6pt
    {\textsuperscript{1}AGI Lab, Westlake University \quad
     \textsuperscript{2}University of Illinois Urbana-Champaign \\
     \textsuperscript{3}Nanyang Technological University\par}
    \vskip 4pt
    {\small\textsuperscript{$\dagger$}Corresponding author. \quad
     \textsuperscript{*}Equal contribution.\par}
    \vskip 5pt
    {\small Project page: \url{https://ashleyxyz.github.io/Slot-3R/}\par}
  \end{center}
  \vskip 6pt
  \renewenvironment{abstract}{\centerline{\large\bfseries Abstract}\vskip 7pt\itshape\noindent\ignorespaces}{\par}
  \begin{abstract}
  Streaming 3D reconstruction must preserve evidence from each frame while
  processing an expanding scene online.  Spatial memory is a natural fit because
  it organizes history by reconstructed 3D location.  Yet Point3R uses spatial
  proximity both to associate a new observation with an existing memory entry
  and to decide whether to fuse it, conflating co-location with state identity.
  Because pointers summarize image patches, nearby pointers may encode distinct
  surfaces, viewpoints, or visibility conditions; averaging them can destroy
  complementary evidence before later frames disambiguate it.  We argue that
  location should determine address, not whether observations must merge.
  \method{} realizes this principle as a training-free, set-associative retrofit
  that lets multiple states coexist at a shared address while keeping the
  pretrained Point3R backbone frozen.  A bounded sparse readout further
  decouples persistent storage from per-frame decoder access.  At 300--500
  sampled frames, \method{} reduces Point3R's point-cloud accuracy error (Acc)
  by 57.1\%--63.1\% on 7Scenes and 64.0\%--72.0\% on NeuralRGBD, lowers
  Sim(3)-aligned absolute trajectory error (ATE) on all three pose benchmarks,
  and remains competitive on video-depth estimation.  It completes all evaluated
  settings from 600 to 1000 sampled frames at about 19 FPS under the same
  protocol, whereas Point3R and InfiniteVGGT run out of memory at 800 frames and
  beyond.

\end{abstract}

\end{tcolorbox}
\vskip 0.12in

\begingroup
\setlength{\intextsep}{6pt}
\begin{figure}[H]
  \centering
  \includegraphics[width=0.90\textwidth]{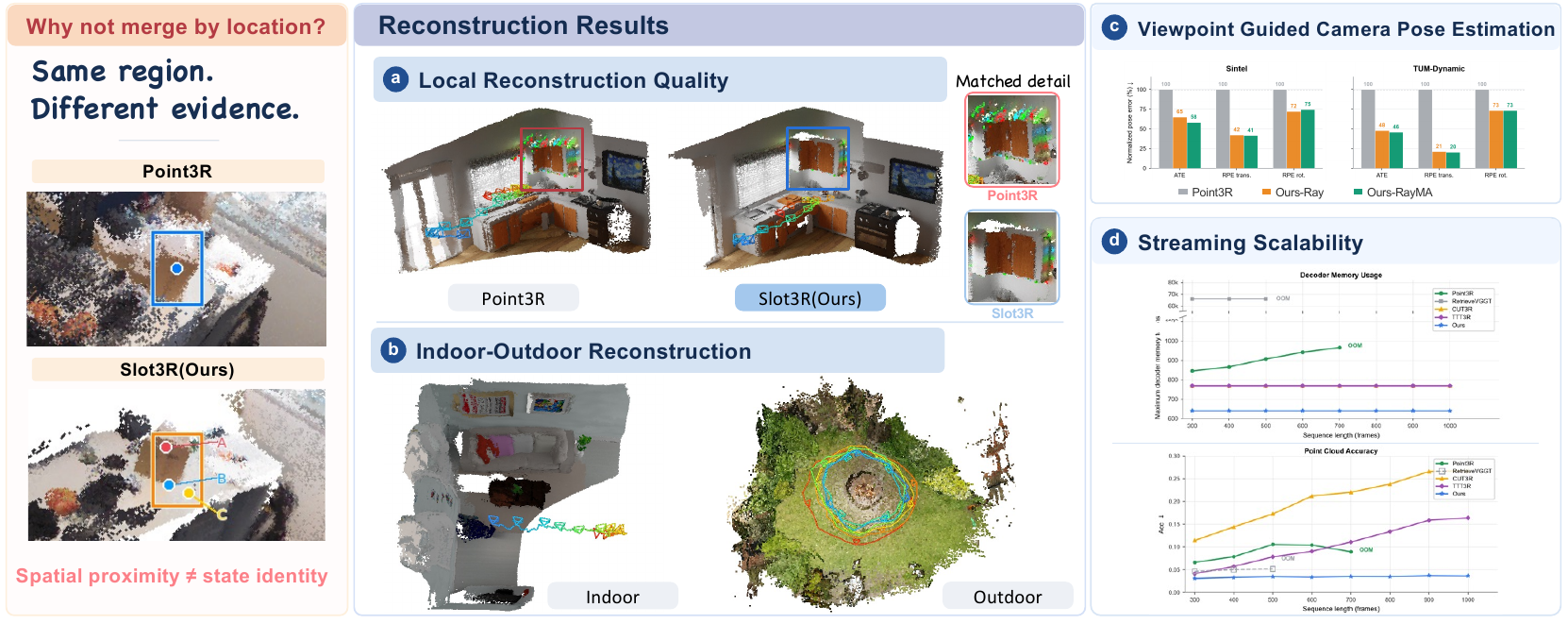}
  \caption{Reconstruction quality, viewpoint-guided pose estimation, and
  streaming scalability.  (a) Local point-cloud comparison with Point3R.
  (b) Indoor and outdoor reconstructions produced by Ours.  (c) Relative to
  Point3R, both VPC variants reduce all three pose errors on Sintel and
  TUM-Dynamic; Ours-VPC-A lowers ATE by 41.8\%/53.1\% and translation RPE by
  58.6\%/79.6\%, respectively.  (d) The 640-token readout keeps decoder-visible
  memory bounded: Slot3R completes 1000-frame streams with Acc \(\leq0.062\),
  while Point3R fails with OOM at 800 frames.}
  \label{fig:teaser}
\end{figure}
\endgroup

\section{Introduction}
\label{sec:intro}

Recovering 3D scene geometry and camera motion from visual observations is a
foundational capability for visual SLAM, robotic navigation, augmented
reality, and embodied intelligence. Classical Structure-from-Motion and
multi-view stereo pipelines rely on feature matching, triangulation, and
iterative optimization
\citep{snavely2006photo,agarwal2009rome,schonberger2016sfm,furukawa2010accurate,schonberger2016mvs}.
Recent feed-forward models instead regress pointmaps, depth, and camera
parameters from multiple views in a single forward pass
\citep{wang2024dust3r,yang2025fast3r,wang2025pi3,wang2025vggt}. Although this
shift substantially reduces reconstruction time, most such models require the
complete image collection before inference. Real-world visual systems instead
receive observations sequentially: future frames are unavailable, each frame
must be incorporated online, and the scene representation may need to expand
throughout operation. Streaming 3D reconstruction must therefore preserve and
organize accumulated visual evidence without repeatedly processing the full
observation history.

An attractive route is to anchor long-term memory directly to the
reconstructed 3D scene \citep{wang2024spann3r,wu2025point3r}. Spatially indexed
memory associates historical
features with reconstructed 3D locations: new regions can allocate memory,
whereas revisited regions can retrieve and update the states stored there.
Compared with fixed-length recurrent states, spatial memory avoids repeatedly
compressing an expanding scene into a limited token set
\citep{wang2025cut3r,ren2026stateupdate}. Causal key--value caches instead retain
history in temporal order, and their storage and attention costs grow with the
stream unless tokens are compressed, selected, or discarded
\citep{zhuo2026streamvggt,yuan2026infinitevggt,zou2026retrievevggt}.
Spatially indexed memory explicitly associates evidence with the reconstructed
3D regions it describes. It therefore organizes history by scene coverage
rather than frame order, making it a promising basis for long-term streaming
reconstruction.
A representative spatial-memory approach is Point3R\citep{wu2025point3r}, which maintains a persistent set of memory entries anchored to reconstructed 3D locations. As each new frame arrives, the model converts its visual evidence into spatially localized memory states and uses nearby stored entries to incorporate information from previously observed regions. This design provides a natural way to organize long-term history by scene location rather than by frame order.

However, an ambiguity arises when spatial proximity is used as a proxy for state identity. In learned spatial memory, a pointer location is a coarse geometric summary of the visual region it represents; for example, Point3R~\citep{wu2025point3r} derives it by averaging the predicted 3D points within an image patch. Consequently, observations of different surfaces, viewpoints, or visibility conditions may occupy nearby 3D locations while carrying distinct and complementary evidence. Merging them solely by proximity can prematurely and irreversibly collapse useful information, producing ambiguous memory states that affect subsequent predictions and updates. This motivates us to decouple spatial association from state consolidation: observations should be organized by location while remaining distinct until their content provides sufficient evidence for fusion.

We introduce \method{}, a spatial pointer memory framework with \emph{set-associative spatial storage}. The core idea is to use spatial proximity only to organize observations into local memory regions, while relying on their feature agreement to determine whether they should be consolidated or preserved as distinct states. Specifically, we quantize 3D space with a spherical hash, where each spatial bucket maintains up to \(K\) independent position--feature slots. For each incoming candidate, a frame-wise confidence filter first removes unreliable observations, after which feature similarity determines its memory update: a sufficiently similar and reliable candidate is fused with its matched slot, while a distinct candidate is stored separately in a free slot or replaces the least reliable one when the bucket is full. In this way, \method{} preserves complementary evidence within the same spatial region instead of prematurely collapsing it based on spatial proximity alone.

A second challenge lies in scaling spatial memory to long streams: as the camera continuously explores new regions, more spatial buckets become occupied and the persistent memory grows with scene coverage, while each incoming frame typically requires only a small fraction of this history. This motivates us to decouple persistent storage from per-frame memory access, allowing the system to preserve an expanding scene history without exposing the entire memory to the decoder at every step. \method{} retrieves relevant local evidence from neighboring spatial buckets using a spatial prior, and complements it with a small set of uniformly sampled global anchors for coarse nonlocal context. Only a fixed number of these selected tokens are passed to the pointer--image interaction module. As a result, \method{} preserves long-term spatial evidence as the scene expands while keeping memory-dependent decoder cost bounded regardless of the total memory size.

\method{} is entirely training-free and can be directly applied to a pretrained Point3R model without modifying or fine-tuning its network parameters.
\method{} sets a new state of the art on 7Scenes and remains competitive with
state-of-the-art methods on NeuralRGBD.
Across 300--500 sampled-frame evaluations, \method{} reduces Point3R's Acc by
57.1\%--63.1\% on 7Scenes and 64.0\%--72.0\% on NeuralRGBD, while increasing
average throughput by 14.3\% and 5.6\%, respectively. It also improves
Sim(3)-aligned ATE on all three pose benchmarks, remains competitive on video
depth, and completes all evaluated streams of up to 1000 sampled frames.

Our contributions are summarized as follows:
\begin{itemize}
  \item We reveal that conflating spatial proximity with state identity can cause premature information collapse in streaming spatial memory.

  \item We introduce \method{}, a training-free set-associative spatial memory that preserves distinct local states while selectively consolidating consistent observations.

  \item We develop a bounded sparse spatial readout that decouples allocated persistent storage from per-frame access cost.

  \item  Without any additional training, \method{} sets a new state of the art for long-sequence point-cloud reconstruction on 7Scenes, while scaling robustly to streams of up to 1000 frames.
\end{itemize}

\section{Related Work}
\label{sec:related}

\subsection{Offline Feed-Forward Visual Geometry}

Feed-forward visual geometry replaces the correspondence-and-optimization chain
of classical reconstruction with direct regression of geometry in a shared
coordinate frame \citep{wang2024dust3r}.  Later offline models scale this up:
many views in a single pass, joint prediction of cameras, depth, pointmaps, and
tracks, and reference-free, permutation-equivariant architectures
\citep{yang2025fast3r,wang2025vggt,wang2025pi3,chen2026hdvggt}.  Further work
cuts the cost on large collections by chunking, partitioning, pruning tokens,
and training scene states at test time \citep{deng2026vggtlong,park2026davggt,
shen2025fastvggt,xie2026scal3r,jin2026zipmap}.  Similar formulations extend to
dynamic and 4D scenes \citep{zhang2025monst3r,zhang2025d4rt,lin2026sm4rt}.  Such
models provide geometric priors for downstream 3D tasks.  Most require full
collections in offline or bidirectional modes, although ZipMap also supports
sequential inference \citep{jin2026zipmap}.  Online methods instead provide
immediate per-frame estimates with stream-scalable memory and computation.

\subsection{Streaming Reconstruction}

Streaming reconstruction preserves past observations in two memory forms.
One compresses the stream into a bounded latent state,
rewriting a fixed token pool across frames \citep{wang2025cut3r}.  Per-frame
cost is constant; later work refines the write rule or uses
sequence models and test-time-trained states
\citep{chen2025ttt3r,dong2026memix,ren2026stateupdate,jin2026filt3r,
deng2026mambavggt}.  LONG3R combines recurrent memory gating with adaptive 3D
spatio-temporal management \citep{chen2025long3r}.  The second causalizes offline
attention, caching earlier-frame keys and values as memory
\citep{zhuo2026streamvggt}.  Frame-level evidence then survives uncompressed,
but the cache grows with the stream and is bounded by windows, hierarchies,
or scoring, merging, evicting, and retrieving tokens under a
fixed budget \citep{li2025wint3r,chen2026gct,chen2026ghost,
xu2026framevggt,liu2026streamcachevggt,fang2026incvggt,wang2026stac,
cheng2026horizonstream}.  Both designs struggle with long streams.  Fixed
capacity may overwrite early evidence in unbounded scenes
\citep{ren2026stateupdate}.  Bounded caches retain history using attention-,
confidence-, or geometry-aware criteria
\citep{chen2026ghost,zou2026retrievevggt}, but remain token-centric rather than
maintaining distinct states at each 3D location.  Geometry-informed external
memory offers another route: Spann3R retrieves by feature affinity,
with geometry encoded in its values \citep{wang2024spann3r}.  Point3R instead
indexes position--feature pointers by 3D location and averages nearby pointers
into one state per region \citep{wu2025point3r}.  Concurrent work adds viewing
direction and retain-or-replace updates \citep{li2026rayaware}.  Our set-associative
memory preserves multiple states per address; viewpoint-guided variants use a
separate pose-conditioning bank.

\section{Method}
\label{sec:method}

\begin{figure}[t]
  \centering
  \includegraphics[width=\linewidth]{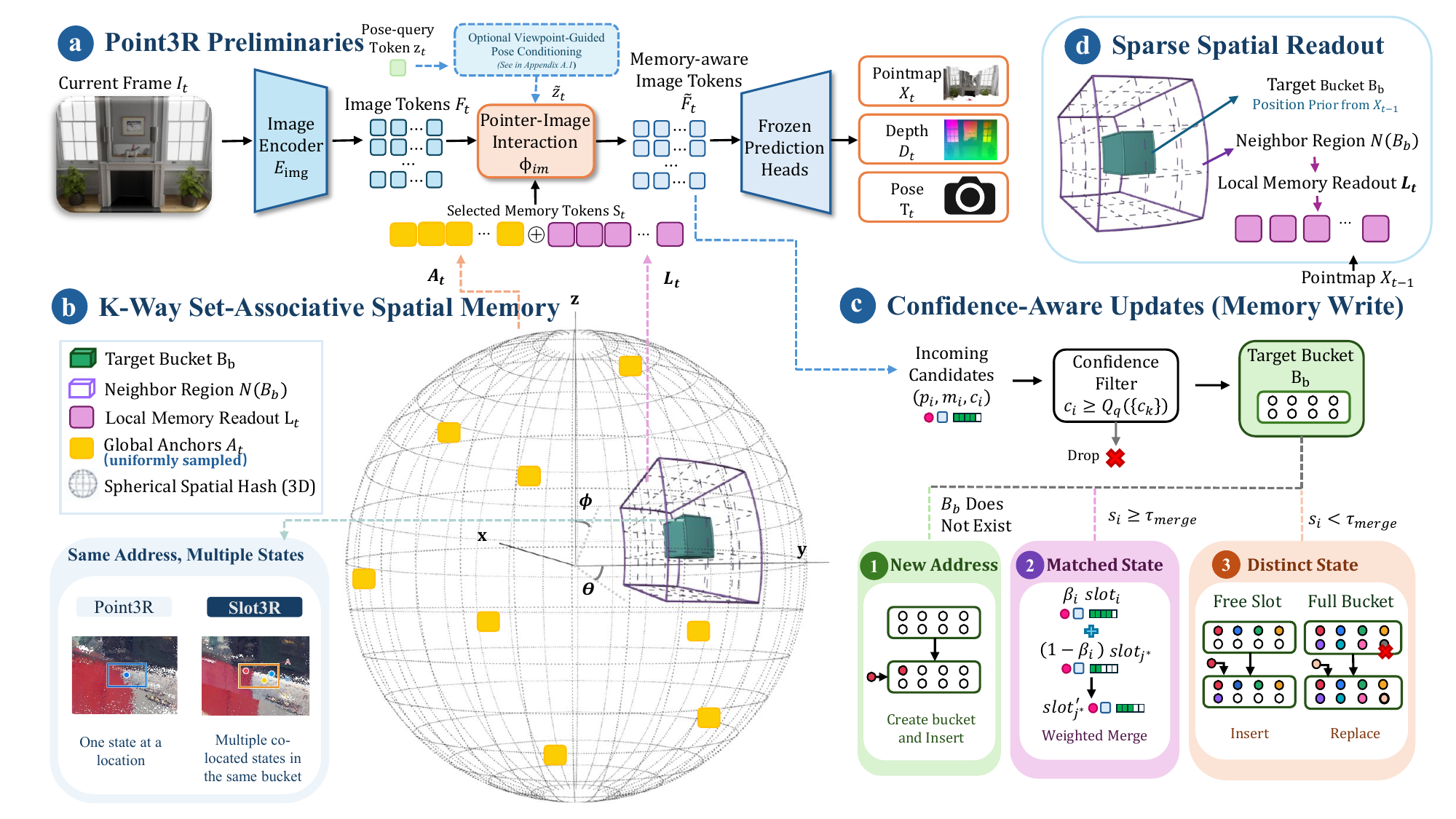}
  \caption{Overview of the core \method{} pipeline.  (a) The frozen Point3R
  backbone interacts image tokens with selected memory states and predicts
  geometry and camera pose (Sec.~\ref{sec:point3r-preliminaries}).  (b) K-way
  set-associative spatial memory retains multiple states at each discrete 3D
  address (Sec.~\ref{sec:set-associative-memory}).  (c) Confidence and feature
  similarity govern the memory-write path through rejection, insertion, fusion,
  and replacement
  (Sec.~\ref{sec:confidence-aware-update}).  (d)
  Previous-frame spatial priors retrieve neighboring buckets; the resulting
  local states are combined with uniformly sampled global anchors under a
  fixed readout budget (Sec.~\ref{sec:sparse-readout}).}
  \label{fig:pipeline}
\end{figure}

We introduce \method{}, a training-free streaming 3D reconstruction method
with a set-associative spatial memory that retains multiple feature states at
each spatial address.  We formulate the problem and review Point3R's
memory in Sec.~\ref{sec:point3r-preliminaries}, followed by our memory
organization in Sec.~\ref{sec:set-associative-memory} and its confidence-aware
update in Sec.~\ref{sec:confidence-aware-update}, and finally the sparse readout
with global anchors in Sec.~\ref{sec:sparse-readout}.  We describe optional
viewpoint-guided pose conditioning in Sec.~\ref{sec:vpc}.  Fig.~\ref{fig:pipeline}
shows the pipeline.

\subsection{Problem Setup and Preliminaries}
\label{sec:point3r-preliminaries}

Streaming 3D reconstruction takes an ordered RGB sequence
\(\mathcal{I}=\{I_t\}_{t=1}^{T}\) and predicts a camera pose \(T_t\), a depth
map \(D_t\), and a pointmap \(X_t\) for each frame as it arrives.  At time \(t\),
the model can access only the prefix \(\mathcal{I}_{1:t}\).  It
therefore maintains a memory \(\mathcal{M}_t\) that summarizes the observations
processed so far.  The central challenge is to organize and update
\(\mathcal{M}_t\) as the stream progresses.

As summarized in Fig.~\ref{fig:pipeline}(a), our method builds on Point3R's
spatial pointer memory~\citep{wu2025point3r}.  In Point3R, each incoming frame
is encoded into image tokens that interact with the existing pointer memory to
predict geometry and camera pose, after which the memory-aware features and
predicted geometry are encoded into new spatial pointers and fused back into
memory.  Given the current image \(I_t\) and the previous memory
\(\mathcal{M}_{t-1}\), this process is written as:
\begin{equation}
  \begin{aligned}
    F_t &= \operatorname{Encode}(I_t), \qquad
    \widetilde{F}_t = \operatorname{Read}(F_t,\mathcal{M}_{t-1}), \qquad
    (X_t,D_t,T_t) = \operatorname{Predict}(\widetilde{F}_t),\\
    \mathcal{C}_t &= \operatorname{Pointerize}
    (\widetilde{F}_t,X_t,D_t,T_t), \qquad
    \mathcal{M}_t = \operatorname{Fuse}(\mathcal{M}_{t-1},\mathcal{C}_t).
  \end{aligned}
  \label{eq:point3r-pipeline}
\end{equation}
Here \(F_t\) and \(\widetilde{F}_t\) are the image and memory-aware tokens,
respectively.  The pointer candidates are
\(\mathcal{C}_t=\{(p_i,m_i)\}_{i=1}^{N_t}\), where
\(N_t\) is the number of candidates, and \(p_i\in\mathbb{R}^{3}\) and \(m_i\)
denote a 3D position and its associated memory feature.  However,
Point3R averages spatially nearby pointer candidates, even when they may
represent distinct surfaces or observations.  To address this, we modify its
inference-time memory to decouple spatial storage from feature fusion.  Each
pointer is assigned to a spatial address based on its 3D position, but this
assignment alone does not determine whether it is fused with a feature state.
Each address can maintain up to \(K\) distinct feature states.

\subsection{Set-Associative Spatial Memory}
\label{sec:set-associative-memory}

Specifically, to retain different feature states within the same spatial region, we organize
the persistent memory \(\mathcal{M}_t\) as a set-associative table over a
discretization of 3D space, as illustrated in Fig.~\ref{fig:pipeline}(b).
An addressing function \(h(\cdot)\) then maps each
candidate position \(p_i\) to a discrete address \(b_i\mathrel{:=}h(p_i)\).
Each address indexes a bucket \(\mathcal{B}_{b_i}\) containing
\(n_{b_i}\leq K\) slots, where slot \(j\) stores
\((p_{b_i,j},m_{b_i,j},c_{b_i,j})\), denoting its 3D position, memory feature,
and confidence, respectively.  In Fig.~\ref{fig:pipeline}(b), the single
blue marker for Point3R represents the memory state produced by averaging
nearby pointers at the highlighted location.  The differently colored markers
for Slot3R represent appearance-distinct states retained in separate slots of
the same spatial bucket, separating a shared spatial address from feature-state
identity while preserving up to \(K\) local alternatives.

Specifically, to assign each pointer candidate to a spatial bucket and retrieve that bucket
efficiently, we map its 3D position to a discrete address on a quantized
spherical grid.  Given \(p_i=(x_i,y_i,z_i)\), we first express its position by
its azimuth, elevation, and log-radius:
\begin{equation}
  \theta_i=\operatorname{atan2}(y_i,x_i),\qquad
  \phi_i=\operatorname{atan2}\!\left(z_i,\sqrt{x_i^2+y_i^2}\right),\qquad
  r_i=\log\!\left(\max(\lVert p_i\rVert_2,\rho_{\min})\right).
  \label{eq:spherical-coordinates}
\end{equation}
We partition these three dimensions into \(B_\theta\), \(B_\phi\), and
\(B_r\) bins.  Their bin widths are
\(\Delta_\theta=2\pi/B_\theta\), \(\Delta_\phi=\pi/B_\phi\), and
\(\Delta_r=(r_{\max}-r_{\min})/B_r\), where
\(r_{\min}=\log\rho_{\min}\) and \(r_{\max}=\log\rho_{\max}\).  We then
quantize the three coordinates into integer bin indices:
\begin{equation}
  \hat{\theta}_i=\left\lfloor(\theta_i+\pi)/\Delta_\theta\right\rfloor,
  \quad
  \hat{\phi}_i=\left\lfloor(\phi_i+\pi/2)/\Delta_\phi\right\rfloor,
  \quad
  \hat{r}_i=\left\lfloor(r_i-r_{\min})/\Delta_r\right\rfloor.
  \label{eq:hash-quantization}
\end{equation}
After clipping each index to its valid range, we concatenate the three indices
to obtain the address:
\begin{equation}
  b_i=h(p_i)=
  \operatorname{Concat}
  \left(\hat{\theta}_i,\hat{\phi}_i,\hat{r}_i\right).
  \label{eq:spatial-hash}
\end{equation}
The address \(b_i\) indexes its corresponding bucket
\(\mathcal{B}_{b_i}\), providing direct access to the slots associated with
that spatial region during memory updates.

\subsection{Confidence-Aware Memory Update}
\label{sec:confidence-aware-update}

Figure~\ref{fig:pipeline}(c) summarizes the confidence-aware write path.
Prediction quality varies across patches, so we first discard candidates
below the frame-wise \(q\)-quantile confidence threshold.  A candidate mapped
to a new bucket is inserted directly.  For each candidate mapped to an existing
bucket, we identify the most similar slot \(j^*\) and the least confident slot
\(j_{\min}\).  If the maximum cosine similarity exceeds
\(\tau_{\mathrm{merge}}\) and the candidate confidence is at least that of the
matched slot, we update \(j^*\) through confidence-weighted fusion with
coefficient \(\beta_i\).  If the similarity is below
\(\tau_{\mathrm{merge}}\), the candidate is admitted only if its confidence is
at least that of \(j_{\min}\); it then occupies a free slot or replaces the
lowest-confidence slot after grouped fusion if the bucket is full.
All remaining candidates are rejected.
In this way, confidence determines whether memory is
modified, while feature similarity determines how it is modified.  Each update
compares at most \(K\) slots and costs \(O(K)\), independent of the total memory
size.  The confidence-based definition of \(\beta_i\) and the complete
slot-level update are given in Appendix~\ref{sec:app-conf-aware-update} and
Algorithm~\ref{alg:conf-aware-fusion}.

\subsection{Sparse Spatial Readout with Global Anchors}
\label{sec:sparse-readout}

The number of occupied memory buckets grows as the camera explores new regions,
but each incoming frame observes only a small part of the scene.  We therefore
retrieve memory at two scales: local neighborhoods provide nearby geometric
evidence, while a small set of global anchors provides coarse context from the
rest of the scene.  We combine the two sets and limit the number of memory
tokens passed to the decoder with a fixed readout budget.

\textbf{Local spatial readout.}
To retrieve local evidence for each image token, we first estimate where its
current patch lies in the spatial memory.  Following Point3R's 3D hierarchical
positional encoding~\citep{wu2025point3r}, we assign each current image token
the position pooled from the same image-patch region of the previous-frame
global pointmap as an approximate spatial prior.  Consecutive frames typically
overlap, so same-index patch regions often observe nearby scene content,
although no exact correspondence is assumed.  Slot3R additionally uses this prior
for sparse retrieval because the current geometry is available only after
pointer--image interaction.  Let \(\bar p_{t,i}\) denote the prior for the
\(i\)-th current image token and \(b_{t,i}=h(\bar p_{t,i})\) its spatial hash
address, written as
\(b_{t,i}=(b^\theta_{t,i},b^\phi_{t,i},b^r_{t,i})\).  Rather than query the
entire table, we retrieve buckets within a neighborhood of this address.  With
\(d\) controlling the neighborhood extent and
\(\delta_\theta,\delta_\phi,\delta_r\in\mathbb{Z}\cap[-d,d]\), we define:
\begin{equation}
  \mathcal{N}_{d}(b_{t,i}) =
  \bigl\{
  (b^\theta_{t,i}+\delta_\theta,
  b^\phi_{t,i}+\delta_\phi,
  b^r_{t,i}+\delta_r)
  \bigr\}.
  \label{eq:neighborhood}
\end{equation}
During neighborhood expansion, out-of-range indices in all three dimensions
are discarded.

For each token, we collect the position-feature slots from occupied
buckets in \(\mathcal{N}_{d}(b_{t,i})\).  Concatenating slots across
\(N_t\) tokens and removing duplicate references yields the frame-level local
readout \(\mathcal{L}_t\), as illustrated in Fig.~\ref{fig:pipeline}(d).

\textbf{Global anchors and readout budget.}
To retain coarse nonlocal context, we uniformly sample a small set of global
anchors from the full memory, denoted by \(\mathcal{A}_t\).
The final readout \(S_t\) places
\(\mathcal{A}_t\) before the local candidates
\(\mathcal{L}_t\), removes duplicate references, and retains the first
\(K_{\mathrm{read}}\) entries.  Thus, at most \(K_{\mathrm{read}}\) memory tokens
participate in pointer-image interaction, combining local geometric evidence
with global context while bounding attention cost independently of persistent
memory size.

\subsection{Auxiliary Viewpoint-Guided Pose Conditioning}
\label{sec:vpc}

The set-associative memory in Secs.~3.2--3.4 is designed primarily to
preserve spatial evidence for reconstruction.  Camera-pose estimation,
however, can additionally benefit from viewpoint-specific history.  We
therefore maintain an auxiliary viewpoint bank \(\mathcal{R}_t\), adapted from
the viewpoint-conditioned pointer representation and update of
\citet{li2026rayaware}, in parallel with the core Slot3R memory
\(\mathcal{M}_t\). For frame \(t\), nearby entries from \(\mathcal{R}_{t-1}\) are retrieved using
the same spatial priors as Sec.~3.4 and pooled with the pose-query token
\(\mathbf{z}_t\) to obtain a viewpoint-conditioned context \(\mathbf{u}_t\).  An
agreement score \(a_t\in[0,1]\) suppresses unreliable retrievals.  We mix this
context into the pose query as
\begin{equation}
  \widetilde{\mathbf{z}}_t
  =
  (1-w_t)\mathbf{z}_t + w_t\mathbf{u}_t,
  \qquad
  w_t =
  \begin{cases}
    a_t \lambda_{\mathrm{j}} g_t, & \text{Ours-VPC-M},\\[2pt]
    a_t \lambda_{\mathrm{c}}, & \text{Ours-VPC-A},
  \end{cases}
  \label{eq:vpc}
\end{equation}
where \(g_t\in[0,1]\) measures recent rotational-motion change.  Ours-VPC-M
emphasizes the auxiliary context when rotational motion changes
sharply, whereas Ours-VPC-A removes the motion-dependent gate and uses a
stronger constant agreement-gated coefficient.  Both variants remain
training-free and use the same single decoder forward.  Details are
provided in Appendix~\ref{app:vpc}.

\definecolor{rankonebg}{RGB}{244,155,177}
\definecolor{ranktwobg}{RGB}{250,193,133}
\definecolor{rankthreebg}{RGB}{255,228,118}
\providecommand{\mone}[1]{\cellcolor{rankonebg}#1}
\providecommand{\mtwo}[1]{\cellcolor{ranktwobg}#1}
\providecommand{\mthree}[1]{\cellcolor{rankthreebg}#1}
\DeclareRobustCommand{\rankonekey}{\begingroup\setlength{\fboxsep}{0.5pt}\raisebox{0pt}[0pt][0pt]{\colorbox{rankonebg}{Pink}}\endgroup}
\DeclareRobustCommand{\ranktwokey}{\begingroup\setlength{\fboxsep}{0.5pt}\raisebox{0pt}[0pt][0pt]{\colorbox{ranktwobg}{Orange}}\endgroup}
\DeclareRobustCommand{\rankthreekey}{\begingroup\setlength{\fboxsep}{0.5pt}\raisebox{0pt}[0pt][0pt]{\colorbox{rankthreebg}{Yellow}}\endgroup}
\providecommand{\deltacell}[1]{{\scriptsize #1}}

\begin{figure*}[!t]
  \centering
  \includegraphics[width=\textwidth]{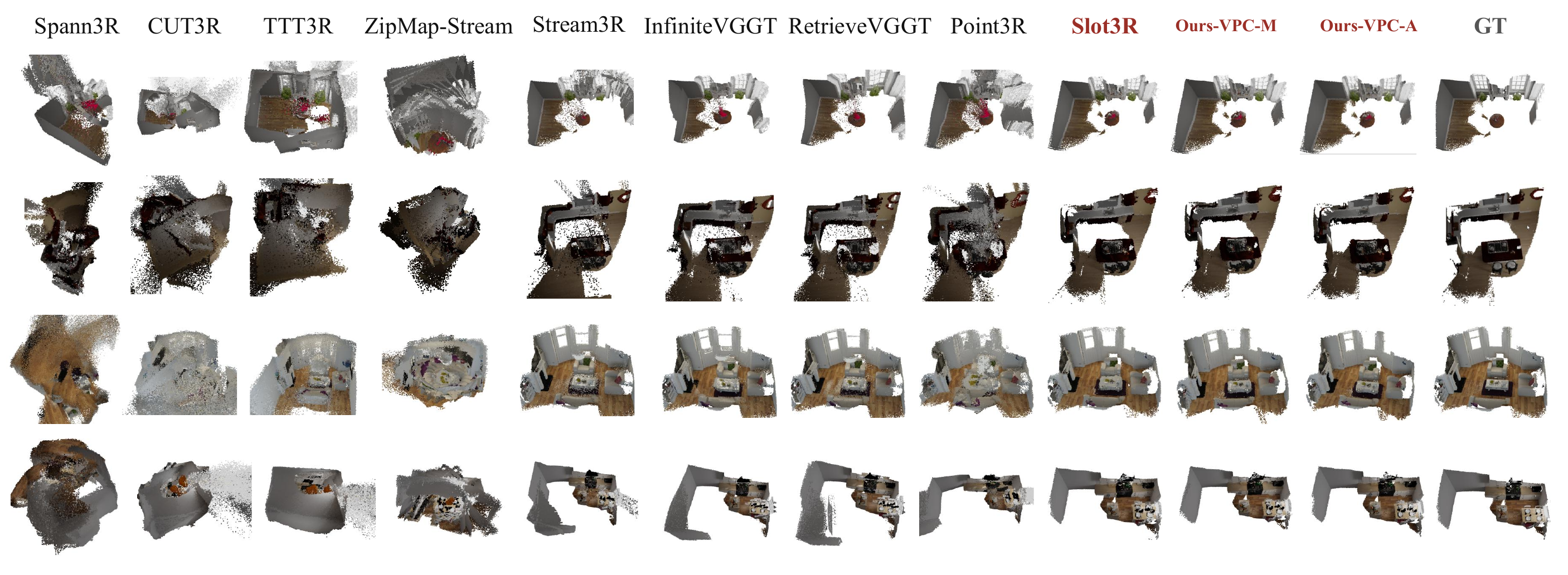}
  \caption{Qualitative point-cloud reconstruction comparison on four
  500-frame NeuralRGBD sequences.  Our variants recover more complete and coherent
  scene geometry than Point3R and more closely approach the ground truth.}
  \label{fig:visual-comparison}
\end{figure*}
\section{Experiments}
\label{sec:experiments}

\begin{table}[t]
  \centering
  \caption{Point-cloud reconstruction on 7Scenes and NeuralRGBD with \(k=2\)
  at 300--500 frames.  \rankonekey, \ranktwokey, and \rankthreekey\ denote the
  top three results per block; ties share and occupy the corresponding ranks.  \(\Delta\) reports the relative
  change from Point3R to Ours (Core), and OOM denotes failure.  Avg. FPS averages
  completed runs.  \(\dagger\) marks reconstruction values reported by
  RetrieveVGGT~\citep{zou2026retrievevggt}; FPS is measured in our setup.}
  \label{tab:pointcloud}
  {\small
  \setlength{\tabcolsep}{2pt}
  \renewcommand{\arraystretch}{0.82}
  \fitwidth{
  \begin{tabular}{lcccccccccc}
    \toprule
    Method &
    \multicolumn{3}{c}{Acc \(\downarrow\)} &
    \multicolumn{3}{c}{Comp \(\downarrow\)} &
    \multicolumn{3}{c}{NC \(\uparrow\)} &
    Avg. FPS \(\uparrow\) \\
    \cmidrule(lr){2-4}\cmidrule(lr){5-7}\cmidrule(lr){8-10}
    & 300 & 400 & 500
    & 300 & 400 & 500
    & 300 & 400 & 500 & \\
    \midrule
    \multicolumn{11}{l}{\textbf{\textit{7Scenes}}} \\
    CUT3R & 0.1362 & 0.1697 & 0.1926 & 0.0744 & 0.1078 & 0.0894 & 0.5513 & 0.5447 & 0.5409 & 13.89 \\
    TTT3R\textsuperscript{\(\dagger\)} & 0.0402 & 0.0502 & 0.0651 & 0.0245 & 0.0263 & 0.0295 & 0.5647 & 0.5575 & 0.5522 & 13.68 \\
    StreamVGGT & OOM & OOM & OOM & OOM & OOM & OOM & OOM & OOM & OOM & OOM \\
    STream3R & 0.0962 & 0.1013 & OOM & 0.0317 & 0.0426 & OOM & 0.5769 & 0.5744 & OOM & 5.82 \\
    InfiniteVGGT\textsuperscript{\(\dagger\)} & 0.0432 & 0.0421 & 0.0398 & 0.0269 & 0.0269 & 0.0236 & 0.5695 & 0.5649 & 0.5617 & 3.63 \\
    RetrieveVGGT & 0.0266 & 0.0276 & 0.0284 & 0.0275 & 0.0292 & 0.0294 & 0.5065 & 0.5063 & 0.5084 & 0.97 \\
    ZipMap-Stream & 0.1096 & 0.1057 & 0.1021 & 0.2644 & 0.3001 & 0.3088 & 0.5223 & 0.5180 & 0.5150 & 5.12 \\
    \hdashline
    \noalign{\vskip 0.3ex}
    Spann3R\textsuperscript{\(\dagger\)} & 0.0765 & 0.0797 & 0.0707 & 0.0278 & 0.0275 & 0.0187 & 0.5464 & 0.5428 & 0.5402 & \mone{41.80} \\
    Point3R & 0.0634 & 0.0638 & 0.0745 & 0.0347 & 0.0308 & 0.0265 & 0.5596 & 0.5541 & 0.5510 & 16.77 \\
    Ours (Core) & \mtwo{0.0262} & \mthree{0.0274} & \mtwo{0.0275} & \mone{0.0154} & \mthree{0.0169} & \mone{0.0158} & \mtwo{0.5790} & \mone{0.5789} & \mthree{0.5790} & \mtwo{19.17} \\
    Ours-VPC-M & \mone{0.0257} & \mone{0.0266} & \mthree{0.0282} & \mtwo{0.0158} & \mone{0.0164} & \mtwo{0.0163} & \mthree{0.5789} & \mtwo{0.5786} & \mone{0.5792} & \mthree{17.54} \\
    Ours-VPC-A & \mthree{0.0264} & \mtwo{0.0269} & \mone{0.0270} & \mthree{0.0159} & \mtwo{0.0167} & \mtwo{0.0163} & \mone{0.5793} & \mtwo{0.5786} & \mtwo{0.5791} & 15.73 \\
    \midrule
    \deltacell{\textit{\(\Delta\) Ours (Core) vs. Point3R}} & \deltacell{\(\downarrow 58.7\%\)} & \deltacell{\(\downarrow 57.1\%\)} & \deltacell{\(\downarrow 63.1\%\)} & \deltacell{\(\downarrow 55.6\%\)} & \deltacell{\(\downarrow 45.1\%\)} & \deltacell{\(\downarrow 40.4\%\)} & \deltacell{\(\uparrow 3.5\%\)} & \deltacell{\(\uparrow 4.5\%\)} & \deltacell{\(\uparrow 5.1\%\)} & \deltacell{\(\uparrow 14.3\%\)} \\
    \midrule
    \multicolumn{11}{l}{\textbf{\textit{NeuralRGBD}}} \\
    CUT3R & 0.2260 & 0.3170 & 0.3471 & 0.0883 & 0.1245 & 0.1539 & 0.5846 & 0.5656 & 0.5539 & 17.51 \\
    TTT3R & 0.1058 & 0.1425 & 0.1666 & 0.0323 & 0.0775 & 0.1027 & 0.6415 & 0.6269 & 0.6221 & 17.40 \\
    StreamVGGT & OOM & OOM & OOM & OOM & OOM & OOM & OOM & OOM & OOM & OOM \\
    STream3R & 0.0664 & 0.0828 & OOM & \mtwo{0.0149} & 0.0211 & OOM & \mone{0.6880} & \mone{0.6935} & OOM & 5.81 \\
    InfiniteVGGT\textsuperscript{\(\dagger\)} & 0.0532 & 0.0651 & 0.0696 & 0.0244 & 0.0348 & 0.0370 & 0.6431 & 0.6503 & 0.6433 & 3.63 \\
    RetrieveVGGT & 0.0438 & \mone{0.0456} & \mone{0.0370} & 0.0248 & 0.0316 & 0.0397 & 0.6166 & 0.6158 & 0.5132 & 1.07 \\
    ZipMap-Stream & 0.1745 & 0.1985 & 0.2141 & 0.2962 & 0.4452 & 0.5305 & 0.5288 & 0.5288 & 0.5291 & 5.01 \\
    \hdashline
    \noalign{\vskip 0.3ex}
    Spann3R\textsuperscript{\(\dagger\)} & 0.1002 & 0.1417 & 0.1829 & 0.0262 & 0.0447 & 0.0487 & 0.5953 & 0.5855 & 0.5794 & \mone{35.37} \\
    Point3R & 0.1132 & 0.1665 & 0.2106 & 0.0377 & 0.0557 & 0.0706 & 0.6116 & 0.6039 & 0.6025 & \mthree{18.01} \\
    Ours (Core) & \mthree{0.0408} & \mthree{0.0571} & \mtwo{0.0589} & 0.0151 & \mtwo{0.0207} & \mtwo{0.0236} & \mtwo{0.6662} & \mtwo{0.6743} & \mone{0.6728} & \mtwo{19.02} \\
    Ours-VPC-M & \mtwo{0.0403} & 0.0579 & \mthree{0.0606} & \mthree{0.0150} & \mtwo{0.0207} & \mone{0.0218} & 0.6630 & \mthree{0.6711} & \mthree{0.6692} & 17.57 \\
    Ours-VPC-A & \mone{0.0401} & \mtwo{0.0561} & 0.0611 & \mone{0.0147} & \mone{0.0206} & \mthree{0.0242} & \mthree{0.6631} & \mthree{0.6711} & \mtwo{0.6699} & 16.25 \\
    \midrule
    \deltacell{\textit{\(\Delta\) Ours (Core) vs. Point3R}} & \deltacell{\(\downarrow 64.0\%\)} & \deltacell{\(\downarrow 65.7\%\)} & \deltacell{\(\downarrow 72.0\%\)} & \deltacell{\(\downarrow 59.9\%\)} & \deltacell{\(\downarrow 62.8\%\)} & \deltacell{\(\downarrow 66.6\%\)} & \deltacell{\(\uparrow 8.9\%\)} & \deltacell{\(\uparrow 11.7\%\)} & \deltacell{\(\uparrow 11.7\%\)} & \deltacell{\(\uparrow 5.6\%\)} \\
    \bottomrule
  \end{tabular}
  }
  }
\end{table}

\begin{table}[t]
  \centering
  \caption{Video depth estimation on ScanNet, Bonn, and KITTI under
  per-sequence alignment and direct metric-scale evaluation.
  \(\delta<1.25\) is reported as a percentage.}
  \label{tab:depth}
  {\scriptsize
  \setlength{\tabcolsep}{2pt}
  \renewcommand{\arraystretch}{0.82}
  \fitwidth{
  \begin{tabular}{llcccccc}
    \toprule
    Method & Alignment &
    \multicolumn{2}{c}{ScanNet} &
    \multicolumn{2}{c}{Bonn} &
    \multicolumn{2}{c}{KITTI} \\
    \cmidrule(lr){3-4}\cmidrule(lr){5-6}\cmidrule(lr){7-8}
    & & AbsRel \(\downarrow\) & \(\delta<1.25\) \(\uparrow\) &
    AbsRel \(\downarrow\) & \(\delta<1.25\) \(\uparrow\) &
    AbsRel \(\downarrow\) & \(\delta<1.25\) \(\uparrow\) \\
    \midrule
    CUT3R & Per-sequence & 0.056 & 97.1 & 0.074 & 94.4 & 0.111 & 88.3 \\
    TTT3R & Per-sequence & 0.049 & \mthree{97.8} & 0.064 & 95.8 & 0.100 & 91.3 \\
    StreamVGGT & Per-sequence & 0.268 & 57.1 & \mthree{0.059} & \mthree{97.2} & 0.173 & 72.2 \\
    STream3R & Per-sequence & \mone{0.037} & \mone{98.9} & \mtwo{0.058} & \mone{97.5} & \mone{0.077} & \mone{95.2} \\
    InfiniteVGGT & Per-sequence & 0.269 & 57.0 & 0.060 & \mtwo{97.3} & 0.172 & 72.4 \\
    RetrieveVGGT & Per-sequence & 0.268 & 57.1 & \mone{0.057} & \mthree{97.2} & 0.175 & 71.9 \\
    ZipMap-Stream & Per-sequence & \mtwo{0.040} & \mtwo{98.6} & 0.073 & 97.1 & 0.110 & 88.9 \\
    \hdashline
    \noalign{\vskip 0.3ex}
    Spann3R & Per-sequence & 0.176 & 73.7 & 0.231 & 72.3 & 0.336 & 51.9 \\
    Point3R & Per-sequence & 0.053 & 97.1 & 0.066 & 95.8 & 0.093 & \mtwo{93.5} \\
    Ours (Core) & Per-sequence & \mthree{0.046} & \mthree{97.8} & 0.063 & 96.2 & \mtwo{0.088} & \mtwo{93.5} \\
    Ours-VPC-M & Per-sequence & \mthree{0.046} & \mthree{97.8} & 0.063 & 96.3 & \mtwo{0.088} & 93.3 \\
    Ours-VPC-A & Per-sequence & \mthree{0.046} & \mthree{97.8} & 0.064 & 96.3 & \mtwo{0.088} & 93.4 \\
    \midrule
    CUT3R & Metric-scale & 0.084 & 94.2 & 0.103 & 88.6 & \mthree{0.127} & \mthree{84.3} \\
    TTT3R & Metric-scale & 0.082 & 95.0 & 0.090 & 94.2 & \mtwo{0.111} & \mtwo{89.1} \\
    StreamVGGT & Metric-scale & 0.490 & 11.0 & 0.644 & 0.3 & 0.969 & 0.0 \\
    STream3R & Metric-scale & 0.547 & 1.8 & \mone{0.065} & \mone{96.7} & \mone{0.084} & \mone{95.3} \\
    InfiniteVGGT & Metric-scale & 0.491 & 11.2 & 0.646 & 0.3 & 0.969 & 0.0 \\
    RetrieveVGGT & Metric-scale & 0.491 & 11.0 & 0.643 & 0.3 & 0.969 & 0.0 \\
    ZipMap-Stream & Metric-scale & 0.573 & 0.0 & 0.648 & 0.6 & 0.925 & 0.0 \\
    \hdashline
    \noalign{\vskip 0.3ex}
    Spann3R & Metric-scale & 0.820 & 0.0 & 0.850 & 0.0 & 0.959 & 0.0 \\
    Point3R & Metric-scale & 0.069 & 95.6 & \mtwo{0.081} & \mtwo{95.8} & 0.169 & 80.5 \\
    Ours (Core) & Metric-scale & \mone{0.063} & \mtwo{96.5} & \mthree{0.087} & 95.7 & 0.179 & 77.7 \\
    Ours-VPC-M & Metric-scale & \mone{0.063} & \mtwo{96.5} & \mthree{0.087} & \mtwo{95.8} & 0.179 & 77.5 \\
    Ours-VPC-A & Metric-scale & \mone{0.063} & \mone{96.6} & \mthree{0.087} & \mtwo{95.8} & 0.179 & 77.8 \\
    \bottomrule
  \end{tabular}
  }
  }
\end{table}

\begin{table}[t]
  \centering
  \caption{Camera pose estimation after Sim(3) alignment on 94 ScanNet scenes,
  14 Sintel sequences, and 8 TUM-Dynamic sequences.}
  \label{tab:pose}
  {\small
  \setlength{\tabcolsep}{2pt}
  \renewcommand{\arraystretch}{0.82}
  \fitwidth{
  \begin{tabular}{lccccccccc}
    \toprule
    Method &
    \multicolumn{3}{c}{ScanNet (Static)} &
    \multicolumn{3}{c}{Sintel} &
    \multicolumn{3}{c}{TUM-Dynamic} \\
    \cmidrule(lr){2-4}\cmidrule(lr){5-7}\cmidrule(lr){8-10}
    & ATE RMSE \(\downarrow\) & RPE trans \(\downarrow\) & RPE rot \(\downarrow\)
    & ATE RMSE \(\downarrow\) & RPE trans \(\downarrow\) & RPE rot \(\downarrow\)
    & ATE RMSE \(\downarrow\) & RPE trans \(\downarrow\) & RPE rot \(\downarrow\) \\
    \midrule
    CUT3R & 0.0960 & 0.0180 & \mtwo{0.4830} & \mone{0.2020} & \mone{0.0560} & \mone{0.5290} & 0.0460 & 0.0120 & \mthree{0.3750} \\
    TTT3R & \mone{0.0640} & \mthree{0.0170} & \mone{0.4590} & \mtwo{0.2050} & 0.0740 & \mtwo{0.5930} & \mone{0.0290} & \mthree{0.0110} & \mtwo{0.3270} \\
    StreamVGGT & 0.1607 & 0.0515 & 3.2820 & 0.2589 & 0.1296 & 1.6673 & 0.0604 & 0.0303 & 2.7908 \\
    STream3R & 0.1191 & 0.0407 & 2.3151 & 0.3721 & 0.0923 & 1.1694 & \mtwo{0.0294} & 0.0125 & \mone{0.3098} \\
    InfiniteVGGT & 0.1618 & 0.0515 & 3.3262 & 0.2589 & 0.1296 & 1.6673 & 0.0607 & 0.0305 & 2.7962 \\
    RetrieveVGGT & 0.1624 & 0.0521 & 3.2814 & 0.2599 & 0.1297 & 1.6721 & 0.0603 & 0.0305 & 2.7915 \\
    ZipMap-Stream & 0.3340 & 0.1010 & 1.7810 & 0.5030 & 0.2380 & 1.1570 & 0.1670 & 0.0700 & 1.5070 \\
    \hdashline
    \noalign{\vskip 0.3ex}
    Spann3R & \mtwo{0.0760} & \mthree{0.0190} & \mthree{0.5090} & 3.2490 & 1.2050 & 2.5020 & 0.0720 & 0.0360 & 2.7670 \\
    Point3R & 0.1120 & 0.0310 & 1.1630 & 0.4360 & 0.1360 & 1.5450 & 0.0640 & 0.0280 & 0.6610 \\
    Ours (Core) & 0.0809 & 0.0322 & 1.4235 & 0.4243 & 0.1279 & 2.0687 & 0.0404 & 0.0260 & 0.7231 \\
    Ours-VPC-M & 0.0817 & \mtwo{0.0165} & 1.0528 & 0.2832 & \mthree{0.0574} & \mthree{1.1154} & 0.0307 & \mtwo{0.0059} & 0.4843 \\
    Ours-VPC-A & \mthree{0.0794} & \mone{0.0162} & 1.1103 & \mthree{0.2539} & \mtwo{0.0563} & 1.1529 & \mthree{0.0300} & \mone{0.0057} & 0.4964 \\
    \bottomrule
  \end{tabular}
  }
  }
\end{table}

\subsection{Experimental Setup}

\textbf{Setup.}
We evaluate dense point-cloud reconstruction, camera trajectories, and video
depth to assess accumulated spatial evidence, trajectory stability, and dense
per-frame prediction, respectively.  We additionally report forwarding speed
across stream lengths to characterize long-sequence efficiency.

\textbf{Baselines.}
We compare with CUT3R \citep{wang2025cut3r}, TTT3R
\citep{chen2025ttt3r}, StreamVGGT \citep{zhuo2026streamvggt}, STream3R
\citep{lan2026stream3r}, InfiniteVGGT \citep{yuan2026infinitevggt},
RetrieveVGGT \citep{zou2026retrievevggt}, Spann3R
\citep{wang2024spann3r}, Point3R \citep{wu2025point3r}, and ZipMap
\citep{jin2026zipmap}.  CUT3R and TTT3R propagate fixed-length recurrent
states, while \emph{ZipMap-Stream} adapts ZipMap's persistent scene state to
sequential inference using unit-size chunks.  StreamVGGT, STream3R,
InfiniteVGGT, and RetrieveVGGT preserve history through causal attention and
cache-based context.  Spann3R and Point3R associate memory with reconstructed
geometry; Point3R is our direct baseline because Ours retains its frozen
backbone and changes only inference-time memory organization and readout.

\textbf{Datasets.}
We evaluate point-cloud reconstruction on 7Scenes
\citep{shotton2013scene} and NeuralRGBD \citep{azinovic2022neuralrgbd}.  Following
state-of-the-art RetrieveVGGT \citep{zou2026retrievevggt}, we use
fixed sequence lengths of 300, 400, and 500 frames and sample every second
frame (\(k=2\)).  Following Point3R
\citep{wu2025point3r}, we evaluate camera pose on ScanNet
\citep{dai2017scannet}, Sintel \citep{butler2012sintel}, and TUM-Dynamic
\citep{sturm2012tum}.  For depth, we retain Bonn
\citep{palazzolo2019refusion} and KITTI
\citep{geiger2012kitti}, following Point3R and CUT3R
\citep{wang2025cut3r}, and evaluate ScanNet.  Methods use the same
frames and metric backend within each dataset block.

\textbf{Metrics.}
We report accuracy (Acc), completeness (Comp), normal
consistency (NC), and average forwarding speed (Avg. FPS), preferring lower
Acc and Comp and higher NC and FPS.  After Sim(3) alignment, pose metrics are
absolute trajectory error RMSE, mean relative translation error, and mean
relative rotation-angle error in degrees; relative errors use adjacent-frame
pairs.  Depth metrics are absolute relative error (AbsRel) and the percentage
of pixels satisfying \(\delta<1.25\), under direct metric-scale evaluation and
per-sequence scale-and-shift alignment.

\textbf{Implementation details.}
Point3R-family variants use released Point3R weights without
fine-tuning.  Unless otherwise stated, our model uses \(K=8\) slots per
bucket, a spherical hash with
\((B_{\theta},B_{\phi},B_r)=(16,8,32)\) bins, radius range
\([\rho_{\min},\rho_{\max}]=[0.05,50]\), and uniform log-radius binning,
local readout radius \(d=1\), merge threshold
\(\tau_{\mathrm{merge}}=0.90\), confidence-drop quantile \(q=0.25\),
\(K_{\mathrm{anchor}}=128\) global anchors, and a readout budget of
\(K_{\mathrm{read}}=640\) memory tokens.  Point3R-family runtime results
use the same fused RoPE3D implementation, so FPS differences reflect memory
organization and readout rather than the PyTorch RoPE3D fallback.  As verified
in Appendix~\ref{sec:app-fused-kernel}, the fused kernel changes runtime while
leaving reconstruction metrics nearly unchanged.
Runtime and OOM comparisons use a single NVIDIA RTX 4090 (24\,GB).  FPS
excludes data loading and metric computation.  The decoder and memory updates
are causal, although the evaluation harness encodes each input sequence before
recurrent decoding.  OOM denotes out-of-memory failure.
Ours (Core) denotes core Slot3R.  Ours-VPC-M adds the auxiliary
viewpoint-conditioned pose readout from Sec.~\ref{sec:vpc} with motion gating;
Ours-VPC-A uses equal interpolation endpoints, yielding agreement-only
conditioning, and refreshes its auxiliary bank every frame.

\subsection{Point-Cloud Reconstruction}

Table~\ref{tab:pointcloud} shows that Ours reduces Point3R's Acc by
\(57.1\%\)--\(63.1\%\) on 7Scenes and \(64.0\%\)--\(72.0\%\) on NeuralRGBD,
while improving NC throughout.  The VPC variants retain comparable quality,
and all three variants complete every setting; STream3R fails at 500 frames,
whereas Spann3R is faster but markedly less accurate.  Figure~\ref{fig:visual-comparison}
shows the qualitative gains over Point3R.

\subsection{Depth Estimation}

Under per-sequence alignment, Ours improves Point3R's AbsRel on all three
datasets while matching or improving \(\delta<1.25\) (Table~\ref{tab:depth}).
The VPC variants remain comparable.  Metric-scale results improve on ScanNet
but not Bonn or KITTI, indicating better relative depth consistency without a
consistent gain in absolute-scale calibration.

\subsection{Camera Pose Estimation}

Ours improves Point3R's ATE on all three benchmarks but not its RPE consistently
(Table~\ref{tab:pose}).  Ours-VPC-A reduces Point3R's ATE by
\(29.1\%\)--\(53.1\%\) and translation RPE by \(47.8\%\)--\(79.7\%\), while
Ours-VPC-M gives comparable gains.  Although CUT3R or TTT3R remains stronger
on several metrics, these results support viewpoint-conditioned pose readout.

\subsection{Ablation Study}

\begin{table}[!htbp]
  \centering
  \caption{Component ablation on nine 200-frame NeuralRGBD sequences
  (\(k=2\)).  Ours uses \(q=0.25\) and a 640-token readout; Final tokens is
  the average sequence-end memory size.}
  \label{tab:ablation}
  {\small
  \setlength{\tabcolsep}{2.5pt}
  \renewcommand{\arraystretch}{0.86}
  \fitwidth{
  \begin{tabular}{lccccc}
    \toprule
    Component & Acc \(\downarrow\) & Comp \(\downarrow\) & NC \(\uparrow\) &
    FPS \(\uparrow\) & Final tokens \(\downarrow\) \\
    \midrule
    Point3R & 0.086 & 0.032 & 0.635 & 17.943 & 836.0 \\
    + Set-associative K-way memory & 0.044 & \textbf{0.017} & 0.676 & 16.346 & 1165.2 \\
    + Sparse local readout (640) & 0.042 & \textbf{0.017} & 0.670 & \textbf{19.411} & 1162.3 \\
    \textbf{+ Filtering and anchors (Ours)} & \textbf{0.039} & 0.018 & \textbf{0.680} & 19.283 & \textbf{694.2} \\
    \bottomrule
  \end{tabular}
  }
  }
\end{table}

\begin{table}[!htbp]
  \centering
  \caption{Spatial-prior ablation on nine 200-frame NeuralRGBD sequences
  (\(k=2\)).  The current-frame alternative adds a preliminary prediction
  pass; Final tokens follows Table~\ref{tab:ablation}.}
  \label{tab:position-prior}
  {\small
  \setlength{\tabcolsep}{3pt}
  \renewcommand{\arraystretch}{0.86}
  \fitwidth{
  \begin{tabular}{lccccc}
    \toprule
    Method & Acc \(\downarrow\) & Comp \(\downarrow\) & NC \(\uparrow\) &
    FPS \(\uparrow\) & Final tokens \(\downarrow\) \\
    \midrule
    Sparse640 (\(q=0.25\)) & \textbf{0.039} & \textbf{0.018} & \textbf{0.680} &
    \textbf{19.28} & \textbf{694.2} \\
    + Predicted 3D position prior & \textbf{0.039} & \textbf{0.018} & 0.676 &
    12.29 & 695.4 \\
    \bottomrule
  \end{tabular}
  }
  }
\end{table}

Table~\ref{tab:ablation} shows that K-way memory improves reconstruction, sparse
readout restores throughput, and filtering with anchors reduces storage while
preserving NC.  Table~\ref{tab:position-prior} finds comparable reconstruction
but much lower FPS with a current-frame prior, favoring Point3R's previous-frame
prior.

\FloatBarrier

\section{Conclusion}
\label{sec:conclusion}

We introduced \method{}, a training-free set-associative spatial memory that
separates spatial addressing from state identity and bounds per-frame decoder
access while keeping the pretrained Point3R backbone frozen.  Across the main
300--500-frame evaluations, core Ours reduces Point3R's point-cloud Acc by
57.1\%--72.0\%, improves normal consistency, and provides 5.6\%--14.3\% higher
average throughput; for depth, it improves per-sequence AbsRel on all three
benchmarks and remains broadly comparable overall, although metric-scale
results are mixed.  Ours-VPC-M and Ours-VPC-A maintain comparable
point-cloud and depth performance while substantially improving camera pose,
with Ours-VPC-A reducing Point3R's Sim(3)-aligned ATE by 29.1\%--53.1\% and
translation RPE by 47.8\%--79.7\% across the three pose benchmarks.  Persistent
storage still grows with scene coverage and depends on fixed memory choices;
future work should study adaptive, globally budgeted memory across backbones,
while our results broadly support separating location from identity and storage
from per-frame access in long-lived 3D perception.

\clearpage
{
  \small
  \bibliographystyle{ieeenat_fullname}
  \bibliography{references}
}

\clearpage
\appendix
\begin{center}
  {\Large\bfseries Appendix}
\end{center}
\begingroup
\small
\setlength{\parskip}{4pt}
\newcommand{\appendixcontentsline}[2]{%
  \noindent\hspace*{#1}%
  \hyperref[#2]{\ref*{#2}\quad\nameref*{#2}}%
  \nobreak\dotfill\pageref{#2}\par}
\noindent\textbf{Contents}\par
\appendixcontentsline{0pt}{sec:appendix}
\appendixcontentsline{1em}{app:vpc}
\appendixcontentsline{1em}{sec:app-extended-pointcloud}
\appendixcontentsline{1em}{sec:app-conf-aware-update}
\appendixcontentsline{1em}{sec:app-bucket-capacity}
\appendixcontentsline{1em}{sec:app-readout-budget}
\appendixcontentsline{1em}{sec:app-hash-type}
\appendixcontentsline{1em}{sec:app-fused-kernel}
\endgroup
\medskip

\section{Additional Ablations and Diagnostics}
\label{sec:appendix}

This appendix reports diagnostic experiments that support the design choices
in the main paper.  The results isolate bucket capacity, readout budget, hash
parameterization, sequence length sensitivity, and
the fused RoPE3D implementation used for fair runtime comparison.

\subsection{Viewpoint-Guided Variant Details}
\label{app:vpc}

\begin{figure}[!htbp]
  \centering
  \includegraphics[width=\linewidth]{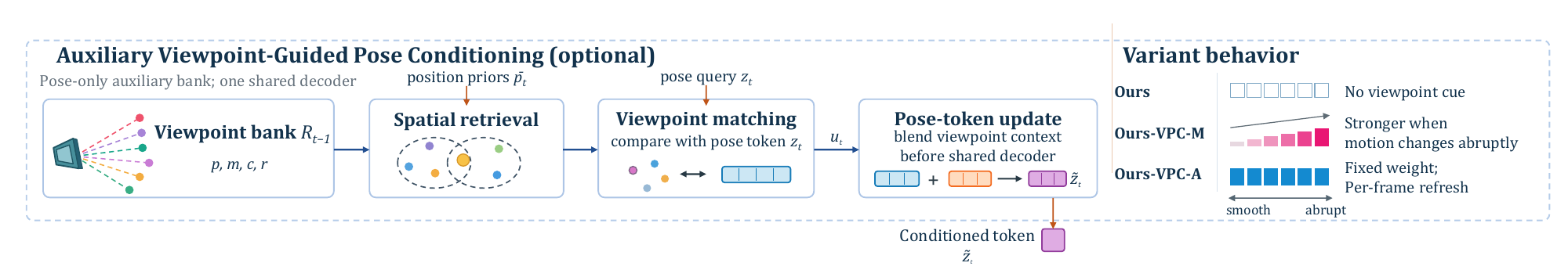}
  \caption{Auxiliary viewpoint-guided pose conditioning.  Spatial priors retrieve
  nearby viewpoint-bank entries, whose features are pooled using the pose query
  to condition its input to the shared decoder.  Ours-VPC-M uses motion- and
  agreement-gated mixing, whereas Ours-VPC-A uses agreement-only mixing with
  per-frame bank refresh.  The auxiliary bank leaves the structure and update
  rule of the core Slot3R memory unchanged, and neither variant requires a
  second decoder forward.}
  \label{fig:app-pose-conditioning}
\end{figure}

This section provides the implementation details of the auxiliary
viewpoint-guided pose conditioning introduced in Sec.~\ref{sec:vpc}.
Figure~\ref{fig:app-pose-conditioning} summarizes the auxiliary pathway and
the two variant-specific conditioning strategies.

The viewpoint bank \(\mathcal{R}_t\) is maintained separately from the core Slot3R
memory \(\mathcal{M}_t\).  Each entry stores a 3D position, a memory feature, a
confidence value, and the viewing direction associated with the observation:
\[
  \mathcal{R}_t
  =
  \left\{
    \left(
      p^r_k,
      m^r_k,
      c^r_k,
      r_k
    \right)
  \right\}.
\]
We adapt the direction-conditioned association and retain-or-replace update of
\citet{li2026rayaware} for this auxiliary bank.  The main Slot3R memory
remains on the set-associative \(K\)-way update described in Sec.~3.3 and
supplies the sparse reconstruction readout of Sec.~3.4.  Thus, the viewpoint
bank enters through pose-token conditioning without changing the structure or
update rule of the core memory.  Because pose and image tokens interact in the
shared decoder, this conditioning can still influence dense predictions and
subsequent memory writes.  Ours-VPC-M updates the auxiliary bank every four
frames, whereas Ours-VPC-A refreshes it every frame.

\paragraph{Viewpoint-bank retrieval and pooling.}
Before processing frame \(t\), we use the spatial position priors
\(\{\bar p_{t,i}\}\) from Sec.~3.4 to retrieve relevant entries from
\(\mathcal{R}_{t-1}\).  We uniformly sample at most 192 position priors and rank
the viewpoint-bank entries by their minimum Euclidean distance to these sampled
positions.  We retain the nearest 128 entries, denoted by \(\mathcal{K}_t\).

Given the current pose-query token \(\mathbf{z}_t\), we compute the cosine
similarity between the query and each retrieved feature,
\[
  s_{t,k}
  =
  \operatorname{sim}
  \left(
    \mathbf{z}_t,
    m^r_k
  \right),
\]
and convert the similarities to attention weights using temperature \(\tau\):
\[
  \alpha_{t,k}
  =
  \frac{
    \exp(s_{t,k}/\tau)
  }{
    \sum_{\ell\in\mathcal{K}_t}
    \exp(s_{t,\ell}/\tau)
  }.
\]
The viewpoint-conditioned feature is then
\[
  \mathbf{u}_t
  =
  \lVert\mathbf{z}_t\rVert_2
  \frac{
    \sum_{k\in\mathcal{K}_t}
    \alpha_{t,k}m^r_k
  }{
    \left\lVert
      \sum_{k\in\mathcal{K}_t}
      \alpha_{t,k}m^r_k
    \right\rVert_2
  }.
\]
The normalization preserves the magnitude of the original pose-query token.

To prevent weakly related viewpoint-bank entries from perturbing the pose query, we
compute an agreement factor from the maximum cosine similarity.  We first
rescale the maximum similarity from \([-1,1]\) to \([0,1]\), clip it to this
interval for numerical stability, and raise it to the fourth power:
\[
  a_t
  =
  \left[
    \operatorname{clip}
    \left(
      \frac{1+\max_{k\in\mathcal{K}_t}s_{t,k}}{2},
      0,1
    \right)
  \right]^4.
\]
We use \(\tau=0.10\) in all reported viewpoint-guided experiments.

\paragraph{Motion gate for Ours-VPC-M.}
Ours-VPC-M additionally modulates the auxiliary context according to the change
in recent rotational motion.  Let \(\mathbf{R}_{t-3}\),
\(\mathbf{R}_{t-2}\), and \(\mathbf{R}_{t-1}\) be the latest three estimated
camera rotations available before processing frame \(t\).  We form the two
relative rotations
\[
  \Delta\mathbf{R}_{t-2}
  =
  \mathbf{R}_{t-3}^{\top}\mathbf{R}_{t-2},
  \qquad
  \Delta\mathbf{R}_{t-1}
  =
  \mathbf{R}_{t-2}^{\top}\mathbf{R}_{t-1}.
\]
The rotational change \(j_t\) is the angular difference between these two
relative rotations.  We linearly map \(j_t\) from \(j_{\mathrm{low}}\) to
\(j_{\mathrm{high}}\) and clip the result to \([0,1]\) to obtain the motion
gate \(g_t\).  We use \(j_{\mathrm{low}}=0.010\) and
\(j_{\mathrm{high}}=0.035\) radians.

The final pose-query mixing is
\[
  \widetilde{\mathbf{z}}_t
  =
  (1-w_t)\mathbf{z}_t+w_t\mathbf{u}_t.
\]
For Ours-VPC-M,
\[
  w_t^{\mathrm{M}}
  =
  a_t\lambda_{\mathrm{j}}g_t,
  \qquad
  \lambda_{\mathrm{j}}=0.025.
\]
The auxiliary context therefore receives negligible weight under smooth
rotational motion and progressively more influence as the change between
consecutive relative rotations increases.

\paragraph{Agreement-only conditioning for Ours-VPC-A.}
Ours-VPC-A uses the same auxiliary bank, spatial retrieval, cosine attention,
and agreement factor, but removes the motion-dependent weighting.  Its mixing
coefficient is
\[
  w_t^{\mathrm{A}}
  =
  a_t\lambda_{\mathrm{c}},
  \qquad
  \lambda_{\mathrm{c}}=0.15.
\]
Hence, motion magnitude does not directly affect the auxiliary residual in
Ours-VPC-A; only the agreement between the pose query and retrieved viewpoint
features controls its frame-dependent strength.  This corresponds to setting
the smooth- and high-jerk endpoints of the underlying interpolation to the same
value, making the interpolation constant.

In both variants, the resulting \(\widetilde{\mathbf{z}}_t\) replaces the
original pose-query token before the shared decoder.  The decoder memory
context still comes from the core Slot3R sparse readout, and no second decoder
forward is introduced.

\subsection{Extended-Sequence Point-Cloud Reconstruction}
\label{sec:app-extended-pointcloud}

Table~\ref{tab:app-extended-pointcloud} evaluates sequences of 600--1000 frames
with \(k=1\), rather than the \(k=2\) sampling used in the main comparison.
This stress test examines whether reconstruction quality and execution remain
stable beyond the sequence lengths reported in Table~\ref{tab:pointcloud}.

\begin{table}[!htbp]
  \centering
  \caption{Extended-sequence point-cloud reconstruction on 7Scenes and
  NeuralRGBD with \(k\)-frame sampling \(k=1\).  ``OOM'' indicates
  out-of-memory failure.  \rankonekey,
  \ranktwokey, and \rankthreekey\ cells mark the best, second-best, and
  third-best results within each dataset--length--metric block and within each
  dataset for Avg. FPS; ties occupy the corresponding ranks.  \(\dagger\)
  RetrieveVGGT exhausts CPU memory because
  its full-history KV repository grows with the input sequence.}
  \label{tab:app-extended-pointcloud}
  {\small
  \setlength{\tabcolsep}{2pt}
  \renewcommand{\arraystretch}{0.82}
  \fitwidth{
  \begin{tabular}{lcccccccccccccccc}
    \toprule
    Method &
    \multicolumn{5}{c}{Acc \(\downarrow\)} &
    \multicolumn{5}{c}{Comp \(\downarrow\)} &
    \multicolumn{5}{c}{NC \(\uparrow\)} &
    Avg. FPS \(\uparrow\) \\
    \cmidrule(lr){2-6}\cmidrule(lr){7-11}\cmidrule(lr){12-16}
    & 600 & 700 & 800 & 900 & 1000
    & 600 & 700 & 800 & 900 & 1000
    & 600 & 700 & 800 & 900 & 1000 & \\
    \midrule
    \multicolumn{17}{l}{7Scenes} \\
    CUT3R & 0.2117 & 0.2200 & 0.2381 & 0.2658 & 0.2674 & 0.1263 & 0.1286 & 0.1279 & 0.1236 & 0.1010 & 0.5620 & 0.5602 & 0.5547 & 0.5502 & 0.5333 & 17.37 \\
    TTT3R & 0.0901 & 0.1106 & 0.1338 & 0.1588 & 0.1638 & 0.0477 & 0.0616 & 0.0670 & 0.0728 & 0.0728 & 0.6217 & 0.6157 & 0.6075 & 0.6018 & 0.6013 & 17.04 \\
    StreamVGGT & OOM & OOM & OOM & OOM & OOM & OOM & OOM & OOM & OOM & OOM & OOM & OOM & OOM & OOM & OOM & OOM \\
    InfiniteVGGT & 0.0533 & 0.0545 & OOM & OOM & OOM & 0.0356 & 0.0378 & OOM & OOM & OOM & \mone{0.6815} & \mone{0.6772} & OOM & OOM & OOM & 3.44 \\
    RetrieveVGGT\(^{\dagger}\) & OOM & OOM & OOM & OOM & OOM & OOM & OOM & OOM & OOM & OOM & OOM & OOM & OOM & OOM & OOM & OOM \\
    ZipMap-Stream & 0.1321 & 0.1236 & 0.1281 & 0.1285 & 0.1233 & 0.2225 & 0.2333 & 0.2345 & 0.2470 & 0.2288 & 0.5671 & 0.5629 & 0.5602 & 0.5543 & 0.5567 & \mthree{17.82} \\
    \hdashline
    \noalign{\vskip 0.3ex}
    Spann3R & 1.0309 & 0.9774 & 0.9379 & 0.9294 & 0.9545 & 1.9878 & 1.9876 & 1.9716 & 1.9669 & 1.9844 & 0.5372 & 0.5246 & 0.5316 & 0.5312 & 0.5226 & \mone{23.85} \\
    Point3R & 0.1039 & 0.0891 & OOM & OOM & OOM & 0.0680 & 0.0432 & OOM & OOM & OOM & 0.6042 & 0.6023 & OOM & OOM & OOM & 17.54 \\
    Ours (Core) & \mtwo{0.0336} & \mthree{0.0351} & \mone{0.0348} & \mthree{0.0369} & \mone{0.0362} & \mthree{0.0231} & \mthree{0.0241} & \mtwo{0.0247} & \mtwo{0.0251} & \mthree{0.0245} & \mthree{0.6663} & \mthree{0.6700} & \mone{0.6703} & \mone{0.6739} & \mone{0.6737} & \mtwo{19.18} \\
    Ours-VPC-M & \mthree{0.0340} & \mone{0.0346} & \mthree{0.0355} & \mone{0.0362} & \mthree{0.0364} & \mone{0.0220} & \mone{0.0231} & \mone{0.0239} & \mone{0.0241} & \mone{0.0237} & 0.6627 & 0.6672 & \mthree{0.6655} & \mthree{0.6679} & \mthree{0.6687} & 17.52 \\
    Ours-VPC-A & \mone{0.0333} & \mtwo{0.0349} & \mtwo{0.0352} & \mone{0.0362} & \mone{0.0362} & \mtwo{0.0228} & \mtwo{0.0240} & \mthree{0.0248} & \mthree{0.0254} & \mtwo{0.0243} & \mtwo{0.6684} & \mtwo{0.6704} & \mtwo{0.6701} & \mtwo{0.6727} & \mtwo{0.6726} & 16.03 \\
    \midrule
    \multicolumn{17}{l}{NeuralRGBD} \\
    CUT3R & 0.3547 & 0.3961 & 0.4217 & 0.4454 & 0.4627 & 0.2166 & 0.2327 & 0.2536 & 0.2590 & 0.2641 & 0.5861 & 0.5884 & 0.5776 & 0.5708 & 0.5782 & 17.76 \\
    TTT3R & 0.1626 & 0.1959 & 0.2463 & 0.2793 & 0.2703 & 0.0621 & 0.0861 & 0.1071 & 0.1263 & 0.1338 & \mtwo{0.6733} & 0.6523 & 0.6380 & 0.6204 & 0.6181 & 17.45 \\
    StreamVGGT & OOM & OOM & OOM & OOM & OOM & OOM & OOM & OOM & OOM & OOM & OOM & OOM & OOM & OOM & OOM & OOM \\
    InfiniteVGGT & 0.0592 & 0.0688 & OOM & OOM & OOM & 0.0364 & 0.0436 & OOM & OOM & OOM & \mone{0.7750} & \mone{0.7725} & OOM & OOM & OOM & 3.44 \\
    RetrieveVGGT\(^{\dagger}\) & OOM & OOM & OOM & OOM & OOM & OOM & OOM & OOM & OOM & OOM & OOM & OOM & OOM & OOM & OOM & OOM \\
    ZipMap-Stream & 0.1780 & 0.1846 & 0.2002 & 0.2150 & 0.2259 & 0.2697 & 0.3160 & 0.3635 & 0.3984 & 0.4293 & 0.5873 & 0.5822 & 0.5773 & 0.5649 & 0.5649 & 17.76 \\
    \hdashline
    \noalign{\vskip 0.3ex}
    Spann3R & 0.6594 & 0.5771 & 0.4579 & 0.4555 & 0.4837 & 2.0183 & 2.0187 & 1.9562 & 1.9908 & 2.0561 & 0.5307 & 0.5246 & 0.5151 & 0.5089 & 0.5025 & \mone{23.99} \\
    Point3R & 0.1650 & 0.1885 & OOM & OOM & OOM & 0.0831 & 0.1004 & OOM & OOM & OOM & 0.6195 & 0.6215 & OOM & OOM & OOM & \mthree{18.19} \\
    Ours (Core) & \mtwo{0.0406} & \mthree{0.0505} & \mtwo{0.0588} & \mone{0.0604} & \mthree{0.0618} & \mthree{0.0147} & \mthree{0.0176} & \mtwo{0.0208} & \mtwo{0.0244} & \mthree{0.0243} & 0.6567 & \mthree{0.6631} & \mtwo{0.6684} & \mtwo{0.6644} & \mthree{0.6643} & \mtwo{18.91} \\
    Ours-VPC-M & \mthree{0.0411} & \mtwo{0.0503} & \mtwo{0.0588} & \mtwo{0.0605} & \mtwo{0.0615} & \mtwo{0.0146} & \mtwo{0.0173} & \mone{0.0206} & \mone{0.0239} & \mtwo{0.0217} & 0.6574 & 0.6616 & \mthree{0.6652} & \mthree{0.6614} & \mtwo{0.6645} & 17.24 \\
    Ours-VPC-A & \mone{0.0405} & \mone{0.0495} & \mone{0.0586} & \mthree{0.0609} & \mone{0.0605} & \mone{0.0144} & \mone{0.0172} & \mthree{0.0209} & \mthree{0.0245} & \mone{0.0216} & \mthree{0.6575} & \mtwo{0.6637} & \mone{0.6693} & \mone{0.6646} & \mone{0.6657} & 16.07 \\
    \bottomrule
  \end{tabular}
  }
  }
\end{table}
All three Slot3R configurations complete the 600--1000-frame settings on both
datasets.  On NeuralRGBD, Ours averages 18.91 FPS; its peak GPU memory at
600, 700, 800, 900, and 1000 frames is 17.6074, 19.8434, 17.3587, 18.6159, and
20.0095 GB, respectively.  The VPC variants sustain 16--17 FPS and stable NC.
Point3R and InfiniteVGGT fail beyond 700 frames, while the recurrent baselines
complete the streams with larger errors.

\FloatBarrier

\subsection{Confidence-Aware Memory Update}
\label{sec:app-conf-aware-update}

\textbf{Candidate filtering and bucket admission.}
For each \((p_i,m_i)\in\mathcal{C}_t\), we compute its confidence \(c_i\) by
averaging the prediction confidence over the corresponding image patch.  We
retain candidates at or above a frame-wise quantile threshold:
\begin{equation}
  \begin{aligned}
    \mathcal{W}_t = \{(p_i,m_i,c_i) \mid {}&
    (p_i,m_i)\in\mathcal{C}_t,
    &c_i \geq
    \operatorname{Quantile}_q(\{c_k\}_{k=1}^{N_t})\}.
  \end{aligned}
  \label{eq:confidence-filter}
\end{equation}
Here \(q\) is the drop quantile, and \(\mathcal{W}_t\) contains the candidates
retained by this frame-level filter.  The address \(b_i=h(p_i)\) identifies the
destination bucket \(\mathcal{B}_{b_i}\).  A candidate whose destination bucket
does not yet exist is admitted directly.  For an existing bucket, admission
depends on feature similarity and the confidence of the stored slots.

For a bucket, we identify its lowest-confidence slot \(j_{\min}\) and the slot
\(j^*\) whose feature is most similar to the candidate.  Let
\(\operatorname{sim}(a,b)=a^\top b/(\lVert a\rVert_2\lVert b\rVert_2)\)
denote cosine similarity.  We then write:
\begin{equation}
  j_{\min}=\arg\min_{1\leq j\leq n_{b_i}}c_{b_i,j},\qquad
  j^*=\arg\max_{1\leq j\leq n_{b_i}}\operatorname{sim}(m_i,m_{b_i,j}),\qquad
  s_i=\operatorname{sim}(m_i,m_{b_i,j^*}).
  \label{eq:slotselect}
\end{equation}
Here \(j^*\) is the selected slot and \(s_i\) is its similarity to the
candidate.  A similar candidate with \(s_i\geq\tau_{\mathrm{merge}}\) is
admitted only if \(c_i\geq c_{b_i,j^*}\), whereas a dissimilar candidate is
admitted if \(c_i\geq c_{b_i,j_{\min}}\); all other candidates are rejected.
The former condition prevents a weaker observation from modifying a more
reliable matched state, while the latter ensures that a distinct candidate
has confidence at least as high as the weakest state in the pre-update snapshot.

\textbf{Matched-slot fusion.}
For a single admitted candidate with
\(s_i\geq\tau_{\mathrm{merge}}\), confidence-weighted fusion is
\begin{equation}
  \begin{gathered}
    \beta_i = \frac{c_i}{c_i+c_{b_i,j^{*}}+\epsilon}, \qquad
    p_{b_i,j^{*}} \leftarrow (1-\beta_i)p_{b_i,j^{*}}+\beta_i p_i,\\
    m_{b_i,j^{*}} \leftarrow (1-\beta_i)m_{b_i,j^{*}}+\beta_i m_i, \qquad
    c_{b_i,j^{*}} \leftarrow (1-\beta_i)c_{b_i,j^{*}}+\beta_i c_i .
  \end{gathered}
  \label{eq:slotfusion}
\end{equation}
The weight \(\beta_i\) gives more reliable evidence greater influence on the
updated position, feature, and confidence.  The implementation evaluates all
candidates in a frame against the same pre-update memory snapshot.  If multiple
candidates target the same slot, it first computes their confidence-weighted
mean position and feature and their mean confidence, then applies
Eq.~\eqref{eq:slotfusion} once using these aggregate quantities.  The
single-candidate rule above is the corresponding singleton case.

\textbf{Distinct-state storage.}
If no matched slot is available, an admitted candidate is stored as a distinct
state.  Within each frame, candidates assigned to the same bucket can occupy
slots that were free in the pre-update snapshot, up to capacity \(K\).  If the
  snapshot bucket is full, the implementation applies at most one replacement
candidate for that bucket.  After grouped fusion, this candidate replaces the
then-lowest-confidence slot; remaining colliding candidates are rejected.  The
vectorized implementation resolves excess candidates in its bucket-grouping
order rather than assuming a stable input order.  Each candidate compares
against at most \(K\) snapshot slots.

Algorithm~\ref{alg:conf-aware-fusion} summarizes this snapshot-based batched
update.

\begin{algorithm}[!htbp]
  \caption{Snapshot-based confidence-aware memory update}
  \label{alg:conf-aware-fusion}
  \begin{algorithmic}[1]
    \Require Previous memory \(\mathcal{M}_{t-1}\), retained write candidates
    \(\mathcal{W}_t=\{(p_i,m_i,c_i)\}\), hash function \(h(\cdot)\), bucket
    capacity \(K\), merge threshold \(\tau_{\mathrm{merge}}\).
    \Ensure Updated memory \(\mathcal{M}_{t}\).
    \State Snapshot \(\mathcal{M}^{0}\gets\mathcal{M}_{t-1}\)
    \State Initialize merge groups \(\{\mathcal{G}_{b,j}\}\) and distinct groups
    \(\{\mathcal{D}_{b}\}\)
    \ForAll{\((p_i,m_i,c_i)\in\mathcal{W}_t\) \textbf{in parallel}}
    \State \(b_i\gets h(p_i)\); read \(n^0_{b_i}\) and slots from
    \(\mathcal{M}^{0}\)
    \If{\(n^0_{b_i}=0\)}
    \State Append \(i\) to \(\mathcal{D}_{b_i}\)
    \Else
    \State Compute \(j_{\min}\), \(j^*\), and \(s_i\) from
    \(\mathcal{M}^{0}\) using Eq.~\eqref{eq:slotselect}
    \If{\(s_i\geq\tau_{\mathrm{merge}}\) and
    \(c_i\geq c^0_{b_i,j^*}\)}
      \State Append \(i\) to \(\mathcal{G}_{b_i,j^*}\)
    \ElsIf{\(s_i<\tau_{\mathrm{merge}}\) and
    \(c_i\geq c^0_{b_i,j_{\min}}\)}
      \State Append \(i\) to \(\mathcal{D}_{b_i}\)
    \Else
      \State Drop \(i\)
    \EndIf
    \EndIf
    \EndFor
    \ForAll{nonempty merge groups \(\mathcal{G}_{b,j}\)}
    \State Compute confidence-weighted \(\bar p_{b,j}\), \(\bar m_{b,j}\),
    and mean confidence \(\bar c_{b,j}\)
    \State Fuse \((\bar p_{b,j},\bar m_{b,j},\bar c_{b,j})\) once into snapshot
    slot \((b,j)\) using Eq.~\eqref{eq:slotfusion}
    \EndFor
    \ForAll{nonempty distinct groups \(\mathcal{D}_{b}\)}
    \If{\(n^0_b<K\)}
    \State Insert up to \(K-n^0_b\) candidates selected by the vectorized
    bucket grouping into snapshot-free slots
    \Else
    \State Select one candidate from \(\mathcal{D}_{b}\) by the same grouping
    order
    \State Replace the lowest-confidence slot of bucket \(b\) after grouped
    fusion with the selected candidate
    \EndIf
    \EndFor
    \State \Return \(\mathcal{M}_t\)
  \end{algorithmic}
\end{algorithm}

Table~\ref{tab:confidence-drop-ablation} favors Drop-then-Compare,
particularly at 400--500 frames.  Global post-write pruning degrades more
sharply because it reduces spatial coverage.

\begin{table}[!htb]
  \centering
  \caption{Ablation of the ordering and scope of confidence-guided memory
  filtering on NeuralRGBD with \(k=2\) and \(n=9\) sequences.  All variants
  retain the same K-way memory, confidence-aware fusion, and 640-token sparse
  readout.  C--D and D--C denote Compare-then-Drop and Drop-then-Compare,
  respectively.  Avg. FPS is computed over the matched 200--500-frame runs.
  This diagnostic sweep was run separately from the main comparison in
  Table~\ref{tab:pointcloud}.  OOM denotes failure on at least one sequence.}
  \label{tab:confidence-drop-ablation}
  {\scriptsize
  \setlength{\tabcolsep}{1.6pt}
  \renewcommand{\arraystretch}{0.80}
  \fitwidth{
  \begin{tabular}{l*{15}{c}c}
    \toprule
    Filtering policy &
    \multicolumn{5}{c}{Acc \(\downarrow\)} &
    \multicolumn{5}{c}{Comp \(\downarrow\)} &
    \multicolumn{5}{c}{NC \(\uparrow\)} &
    Avg. FPS \(\uparrow\) \\
    \cmidrule(lr){2-6}\cmidrule(lr){7-11}\cmidrule(lr){12-16}
    & 200 & 300 & 400 & 500 & 1000
    & 200 & 300 & 400 & 500 & 1000
    & 200 & 300 & 400 & 500 & 1000 & \\
    \midrule
    C--D, bucket 25\%
    & 0.0440 & 0.0450 & 0.0690 & 0.0686 & OOM
    & 0.0180 & 0.0150 & 0.0227 & 0.0204 & OOM
    & 0.6745 & 0.6620 & 0.6692 & 0.6673 & OOM & \textbf{19.52} \\
    C--D, bucket 12.5\%
    & 0.0390 & 0.0460 & 0.0672 & 0.0708 & OOM
    & 0.0180 & 0.0160 & 0.0227 & 0.0217 & OOM
    & 0.6810 & 0.6675 & 0.6733 & 0.6695 & OOM & 18.99 \\
    C--D, global 10\%
    & 0.0420 & 0.0620 & 0.1109 & 0.0861 & OOM
    & 0.0180 & 0.0190 & 0.0362 & 0.0250 & OOM
    & 0.6670 & 0.6490 & 0.6545 & 0.6554 & OOM & 19.26 \\
    C--D, global 15\%
    & 0.0530 & 0.1430 & 0.1753 & 0.1242 & OOM
    & 0.0200 & 0.0290 & 0.0660 & 0.0362 & OOM
    & 0.6625 & 0.6435 & 0.6446 & 0.6469 & OOM & 19.33 \\
    No dropping
    & 0.0390 & 0.0450 & 0.0617 & 0.0650 & OOM
    & 0.0180 & 0.0160 & 0.0214 & 0.0203 & OOM
    & 0.6795 & 0.6635 & 0.6737 & 0.6736 & OOM & 18.91 \\
    \textbf{D--C, 25\% (Ours)}
    & \textbf{0.0389} & \textbf{0.0408} & \textbf{0.0571} & \textbf{0.0589} & \textbf{0.0631}
    & \textbf{0.0178} & 0.0151 & \textbf{0.0207} & 0.0236 & \textbf{0.0239}
    & 0.6801 & 0.6664 & \textbf{0.6743} & 0.6729 & \textbf{0.6758} & 19.02 \\
    \bottomrule
  \end{tabular}
  }
  }
\end{table}

\begin{samepage}
  \paragraph{Drop-quantile sensitivity.}
  Tables~\ref{tab:q-sensitivity} and~\ref{tab:q-depth-sensitivity} show opposing
  task trends.  Stronger filtering generally benefits point-cloud accuracy and
  memory: at \(q=0.25\), NC is highest, memory is \(23.5\%\) smaller, and FPS is
  \(7.1\%\) higher than at \(q=0\).  Depth favors weaker filtering, and
  \(q=0.25\) outperforms \(q=0.35\) on nine of twelve metrics.  We therefore use
  \(q=0.25\) as a cross-task compromise between reconstruction, memory, and depth.

\end{samepage}

\begin{table}[!htb]
  \centering
  \caption{A separate sensitivity sweep of the pre-write drop quantile \(q\)
  on 500-frame NeuralRGBD reconstruction with \(n=9\) sequences.  All variants
  within this sweep use the same K-way memory and 640-token sparse readout.}
  \label{tab:q-sensitivity}
  {\small
  \setlength{\tabcolsep}{3pt}
  \renewcommand{\arraystretch}{0.84}
  \begin{tabular}{lccccc}
    \toprule
    \(q\) & Acc \(\downarrow\) & Comp \(\downarrow\) & NC \(\uparrow\) &
    FPS \(\uparrow\) & Final tokens \(\downarrow\) \\
    \midrule
    0.00 & 0.0649 & 0.0227 & 0.6713 & 17.949 & 1367.3 \\
    0.05 & 0.0637 & 0.0228 & 0.6678 & 19.140 & 1291.4 \\
    0.10 & 0.0621 & \textbf{0.0211} & 0.6713 & 19.230 & 1210.9 \\
    0.15 & 0.0624 & 0.0220 & 0.6698 & 19.141 & 1163.4 \\
    0.20 & 0.0620 & 0.0220 & 0.6702 & 19.253 & 1101.4 \\
    \textbf{0.25} & 0.0589 & 0.0236 & \textbf{0.6729} & 19.216 & 1046.2 \\
    0.30 & 0.0597 & 0.0228 & 0.6708 & 19.209 & 996.2 \\
    0.35 & \textbf{0.0588} & 0.0231 & 0.6673 & \textbf{19.359} &
    \textbf{937.1} \\
    \bottomrule
  \end{tabular}
  }
\end{table}

\begin{table}[!htb]
  \centering
  \caption{Depth sensitivity to the pre-write drop quantile \(q\) on ScanNet,
  Bonn, and KITTI under metric-scale and per-sequence alignment.  These results
  are from a diagnostic sweep separate from the main depth evaluation in
  Table~\ref{tab:depth}.  \(\delta<1.25\) is reported as a percentage.}
  \label{tab:q-depth-sensitivity}
  {\scriptsize
  \setlength{\tabcolsep}{3pt}
  \renewcommand{\arraystretch}{0.84}
  \fitwidth{
  \begin{tabular}{llcccccc}
    \toprule
    Alignment & \(q\) &
    \multicolumn{2}{c}{ScanNet} &
    \multicolumn{2}{c}{Bonn} &
    \multicolumn{2}{c}{KITTI} \\
    \cmidrule(lr){3-4}\cmidrule(lr){5-6}\cmidrule(lr){7-8}
    & & AbsRel \(\downarrow\) & \(\delta<1.25\) \(\uparrow\) &
    AbsRel \(\downarrow\) & \(\delta<1.25\) \(\uparrow\) &
    AbsRel \(\downarrow\) & \(\delta<1.25\) \(\uparrow\) \\
    \midrule
    Metric-scale & 0.00 & 0.06219 & 96.54 & \textbf{0.08453} & \textbf{96.08} & 0.18206 & 77.19 \\
    Metric-scale & 0.05 & 0.06225 & 96.53 & 0.08534 & 95.98 & 0.18179 & 77.07 \\
    Metric-scale & 0.10 & \textbf{0.06179} & \textbf{96.58} & 0.08514 & 96.04 & 0.18216 & 77.33 \\
    Metric-scale & 0.15 & 0.06251 & 96.45 & 0.08521 & 95.97 & 0.18135 & 77.11 \\
    Metric-scale & 0.20 & 0.06230 & 96.43 & 0.08547 & 96.01 & 0.17915 & \textbf{77.72} \\
    Metric-scale & \textbf{0.25} & 0.06317 & 96.43 & 0.08652 & 95.86 & 0.17966 & 77.37 \\
    Metric-scale & 0.30 & 0.06290 & 96.44 & 0.08727 & 95.78 & 0.17871 & 77.58 \\
    Metric-scale & 0.35 & 0.06349 & 96.24 & 0.08730 & 95.83 & \textbf{0.17833} & 77.47 \\
    \midrule
    Per-sequence & 0.00 & \textbf{0.04455} & \textbf{97.84} & 0.06379 & 96.38 & \textbf{0.08708} & 93.39 \\
    Per-sequence & 0.05 & 0.04495 & 97.80 & 0.06364 & 96.43 & 0.08793 & 93.35 \\
    Per-sequence & 0.10 & 0.04465 & 97.83 & 0.06350 & 96.40 & 0.08738 & 93.42 \\
    Per-sequence & 0.15 & 0.04533 & 97.74 & \textbf{0.06297} & \textbf{96.44} & 0.08806 & 93.28 \\
    Per-sequence & 0.20 & 0.04554 & 97.77 & 0.06298 & 96.43 & 0.08832 & 93.23 \\
    Per-sequence & \textbf{0.25} & 0.04622 & 97.72 & 0.06324 & 96.37 & 0.08789 & \textbf{93.44} \\
    Per-sequence & 0.30 & 0.04585 & 97.80 & 0.06360 & 96.39 & 0.08772 & 93.30 \\
    Per-sequence & 0.35 & 0.04675 & 97.76 & 0.06355 & 96.36 & 0.08815 & 93.31 \\
    \bottomrule
  \end{tabular}
  }
  }
\end{table}

\begin{samepage}
  \subsection{Bucket Capacity}
  \label{sec:app-bucket-capacity}

  The bucket capacity \(K\) denotes the maximum number of distinct states retained
  at one spatial address.  Tables~\ref{tab:app-k} and~\ref{tab:app-k-kf20} vary
  only \(K\) under two 7Scenes settings.  In Table~\ref{tab:app-k}, increasing
  \(K\) from 1 to 2 reduces Acc and Comp by 66.7\% and 66.5\%, respectively,
  showing that a single state is insufficient when local ambiguity is strong.
  The trend is smoother when sampling every 20th frame (\(k=20\)), where Point3R is already competitive;
  nevertheless, \(K=8\) improves its Acc and Comp by 9.0\% and 12.7\% while its
  NC differs by only 0.0009.  Across both settings, \(K=1\) is weakest and gains
  saturate by \(K=8\).  With \(k=20\), \(K=16\) uses 67.1\% more final tokens
  than \(K=8\) without improving reconstruction.  Although \(K=16\) gives the
  best reconstruction metrics in Table~\ref{tab:app-k}, it uses 63.8\% more final
  tokens than \(K=8\) and is slower.  We therefore use \(K=8\) as the default
  balance between quality, local-state capacity, and memory cost.

\end{samepage}

\begin{table}[!htbp]
  \centering
  \caption{Bucket-capacity ablation on 7Scenes with \(k=2\) and 200 sampled
  frames, where \(K\) is the maximum number of slots per spatial bucket.  Avg.
  final token count is averaged across evaluated sequences.}
  \label{tab:app-k}
  {\small
  \setlength{\tabcolsep}{3pt}
  \renewcommand{\arraystretch}{0.86}
  \begin{tabular}{lcccccc}
    \toprule
    Method & \(K\) & Acc \(\downarrow\) & Comp \(\downarrow\) &
    NC \(\uparrow\) & FPS \(\uparrow\) & Avg. final tokens \(\downarrow\) \\
    \midrule
    Point3R & -- & 0.047003 & 0.018037 & 0.578354 & 17.746 & 918.0 \\
    Ours (Core) & 1 & 0.089329 & 0.053222 & 0.569152 & 19.911 & \textbf{219.2} \\
    Ours (Core) & 2 & 0.029775 & 0.017823 & 0.580439 & \textbf{19.935} & 250.9 \\
    Ours (Core) & 4 & 0.028163 & 0.016960 & 0.583035 & 19.816 & 440.4 \\
    Ours (Core) & \textbf{8} & 0.026758 & 0.017059 & 0.584917 & 19.405 & 754.3 \\
    Ours (Core) & 16 & \textbf{0.026092} & \textbf{0.016093} & \textbf{0.585791} & 18.958 & 1235.2 \\
    \bottomrule
  \end{tabular}
  }
\end{table}

\begin{table}[!htbp]
  \centering
  \caption{Bucket-capacity ablation on 7Scenes when sampling every 20th frame
  (\(k=20\)).  Avg. final token count is averaged across evaluated sequences.}
  \label{tab:app-k-kf20}
  {\small
  \setlength{\tabcolsep}{3pt}
  \renewcommand{\arraystretch}{0.86}
  \begin{tabular}{lcccccc}
    \toprule
    Method & \(K\) & Acc \(\downarrow\) & Comp \(\downarrow\) & NC \(\uparrow\) &
    FPS \(\uparrow\) & Avg. final tokens \(\downarrow\) \\
    \midrule
    Point3R & -- & 0.034096 & 0.024413 & \textbf{0.649663} & 13.618 & 1406.9 \\
    Ours (Core) & 1 & 0.056078 & 0.039312 & 0.636682 & 16.914 & \textbf{168.3} \\
    Ours (Core) & 2 & 0.042358 & 0.029619 & 0.641545 & 16.199 & 271.4 \\
    Ours (Core) & 4 & 0.034110 & 0.023624 & 0.646727 & \textbf{17.047} & 482.4 \\
    Ours (Core) & \textbf{8} & \textbf{0.031013} & \textbf{0.021317} & 0.648767 & 16.459 & 847.0 \\
    Ours (Core) & 16 & 0.031309 & 0.021690 & 0.648118 & 16.513 & 1415.4 \\
    \bottomrule
  \end{tabular}
  }
\end{table}

\begin{samepage}
  \subsection{Readout Budget Sensitivity}
  \label{sec:app-readout-budget}

  The readout budget controls the number of memory tokens exposed to the
  decoder, not the size of the stored memory.  Tables
  \ref{tab:app-readout-budget-pointcloud} and
  \ref{tab:app-readout-budget-depth} show that performance saturates at moderate
  budgets.  On NeuralRGBD point-cloud reconstruction, increasing the budget from
  128 to 512 improves Acc from 0.043 to 0.039 and NC from 0.653 to 0.680, indicating
  that very small readouts omit useful spatial evidence.  Beyond 512 tokens, Acc
  remains within 0.038--0.039 and NC within 0.677--0.681, with no monotonic gain
  from exposing more tokens.  Depth accuracy similarly plateaus around 640--768
  tokens: 768 is numerically best, but differs from 640 by only 0.00003 AbsRel and
  0.00030 in \(\delta<1.25\).  We therefore use \(K_{\mathrm{read}}=640\) as a
  balanced budget across reconstruction and depth.

\end{samepage}

\begin{table}[!htbp]
  \centering
  \caption{Sparse-readout budget sensitivity on NeuralRGBD point-cloud
  reconstruction with \(k=2\) and 200 sampled frames.  This diagnostic sweep
  was run separately from the component ablation in
  Table~\ref{tab:ablation}.}
  \label{tab:app-readout-budget-pointcloud}
  {\small
  \setlength{\tabcolsep}{3pt}
  \renewcommand{\arraystretch}{0.86}
  \fitwidth{
  \begin{tabular}{lcccc}
    \toprule
    Sparse budget & Acc \(\downarrow\) & Comp \(\downarrow\) &
    NC \(\uparrow\) & FPS \(\uparrow\) \\
    \midrule
    128  & 0.043 & \textbf{0.017} & 0.653 & 20.000 \\
    256  & 0.041 & \textbf{0.017} & 0.665 & 20.068 \\
    384  & 0.040 & \textbf{0.017} & 0.677 & \textbf{20.312} \\
    512  & 0.039 & 0.018 & 0.680 & 19.756 \\
    640  & \textbf{0.038} & 0.018 & 0.677 & 19.114 \\
    768  & \textbf{0.038} & \textbf{0.017} & 0.678 & 18.925 \\
    912  & \textbf{0.038} & \textbf{0.017} & \textbf{0.681} & 19.259 \\
    1024 & 0.039 & 0.018 & 0.680 & 19.362 \\
    \bottomrule
  \end{tabular}
  }
  }
\end{table}
\begin{table}[!htbp]
  \centering
  \caption{Sparse-readout budget sensitivity for depth estimation.}
  \label{tab:app-readout-budget-depth}
  {\small
  \setlength{\tabcolsep}{3pt}
  \renewcommand{\arraystretch}{0.86}
  \fitwidth{
  \begin{tabular}{lcc}
    \toprule
    Sparse budget & AbsRel \(\downarrow\) & \(\delta<1.25\) \(\uparrow\) \\
    \midrule
    128 & 0.06508 & 0.96185 \\
    256 & 0.06427 & 0.96323 \\
    384 & 0.06453 & 0.96245 \\
    512 & 0.06405 & 0.96303 \\
    640 & 0.06326 & 0.96399 \\
    768 & \textbf{0.06323} & \textbf{0.96429} \\
    912 & 0.06392 & 0.96199 \\
    1024 & 0.06390 & 0.96175 \\
    \bottomrule
  \end{tabular}
  }
  }
\end{table}

\subsection{Spherical Hash versus Voxel Hash}
\label{sec:app-hash-type}

The hash parameterization determines which observations share a bucket and
which buckets are treated as neighbors during readout.  We compare the proposed
azimuth--elevation--log-radius grid with a voxel-style Cartesian quantization
under the same sparse-512 setting on seven diagnostic sequences.  As shown in
Table~\ref{tab:app-hash-type}, the spherical hash reduces Acc from 0.1409 to
0.0491 and Comp from 0.0564 to 0.0280, while increasing NC from 0.6076 to
0.7451.  This result indicates that the addressing geometry, rather than hash
lookup alone, materially affects which local states are grouped and retrieved.
We therefore retain the spherical parameterization in the main method, while
treating alternative grids as an implementation choice rather than a separate
contribution.

\begin{table}[!htbp]
  \centering
  \caption{Hash parameterization ablation for sparse-512 readout.}
  \label{tab:app-hash-type}
  {\small
  \setlength{\tabcolsep}{4pt}
  \renewcommand{\arraystretch}{0.86}
  \begin{tabular}{lccccc}
    \toprule
    Hash & \(n\) & Acc \(\downarrow\) & Comp \(\downarrow\) &
    NC \(\uparrow\) & Acc med. \(\downarrow\) \\
    \midrule
    Spherical & 7 & 0.0491 & 0.0280 & 0.7451 & 0.0318 \\
    Voxel & 7 & 0.1409 & 0.0564 & 0.6076 & 0.0991 \\
    \bottomrule
  \end{tabular}
  }
\end{table}

\subsection{Fused RoPE3D Runtime Control}
\label{sec:app-fused-kernel}

Runtime comparisons are sensitive to the RoPE3D implementation.
Table~\ref{tab:app-fused-kernel-nrgbd} compares Point3R before and after
enabling the fused RoPE3D kernel under the NeuralRGBD diagnostic protocol.
The fused kernel raises Point3R throughput from 4.547 to 14.750 FPS while
leaving reconstruction metrics and memory essentially unchanged.  Under the
same fused implementation, the 640-token sparse readout raises throughput from
16.658 to 18.269 FPS relative to full-memory readout, while NC remains 0.753.
This bounded readout uses up to 128 uniformly sampled global anchors and fills
the remaining 640-token budget with up to 512 additional local tokens after
deduplication.  These controls separate acceleration due to fused RoPE3D from
that due to the local-and-global bounded readout.  We therefore use fused
RoPE3D for all Point3R-family runtime comparisons and do not count it as part
of our method.

\begin{table}[!htbp]
  \centering
  \caption{Effect of fused RoPE3D and sparse readout on NeuralRGBD diagnostic
  runs with approximately 50-frame sequences.  ``Ours (Core) w/o sparse readout''
  uses K-way memory and pre-write filtering with \(q=0.25\), but exposes the
  full stored memory to the decoder; Ours (Core) instead applies a bounded readout of
  at most 640 tokens, comprising up to 128 uniformly sampled global anchors and
  up to 512 additional local tokens after deduplication.}
  \label{tab:app-fused-kernel-nrgbd}
  {\small
  \setlength{\tabcolsep}{3pt}
  \renewcommand{\arraystretch}{0.86}
  \fitwidth{
  \begin{tabular}{lccccc}
    \toprule
    Method & Acc \(\downarrow\) & Comp \(\downarrow\) &
    NC \(\uparrow\) & FPS \(\uparrow\) & Avg. stored tokens \(\downarrow\) \\
    \midrule
    Point3R w/o fused RoPE3D & 0.084 & 0.030 & 0.753 & 4.547 & 1571.0 \\
    Point3R w/ fused RoPE3D & 0.083 & 0.030 & 0.752 & 14.750 & 1572.6 \\
    Ours (Core) w/o sparse readout & \textbf{0.061} & \textbf{0.027} &
    \textbf{0.753} & 16.658 & \textbf{1204.1} \\
    Ours (Core) & 0.063 & 0.029 & \textbf{0.753} & \textbf{18.269} & 1220.1 \\
    \bottomrule
  \end{tabular}
  }
  }
\end{table}

\FloatBarrier

\end{document}